\documentclass[sigconf,table,screen,review=false,anonymous=false]{acmart}

\copyrightyear{2019}
\acmYear{2019}
\setcopyright{acmlicensed}
\acmConference[CIKM '19]{CIKM '19: The 28th ACM International Conference on Information and Knowledge Management}{November 03--07, 2019}{Beijing, China}
\acmPrice{15.00}
\acmDOI{10.1145/1122445.1122456}
\acmISBN{978-1-4503-9999-9/18/06}

\usepackage{soul}
\usepackage{url}
\usepackage[utf8]{inputenc}
\usepackage[skip=0pt]{caption}
\usepackage{amsmath}
\usepackage{algorithmic}
\usepackage{algorithm}
\usepackage{verbatim}
\usepackage{amsfonts}
\usepackage{amssymb}
\usepackage{array,multirow}
\usepackage{booktabs}
\usepackage{graphicx}
\usepackage{xcolor}
\usepackage{todonotes}
\presetkeys{todonotes}{inline}{}
\usepackage{listings}
\usepackage{amsthm}
\usepackage{textcomp}
\usepackage{enumitem}
\renewcommand{\textrightarrow}{$\rightarrow$}

\newcommand{\OurApproach}{QAmp}

\newcommand{\smalltt}[1]{\texttt{\small #1}}
\newcommand{\commentA}[2][.5\linewidth]{%
  \leavevmode\hfill\makebox[#1][l]{$\triangleright$~#2}}
\newcommand{\OurParagraph}[1]{\smallskip\noindent\textbf{#1}}

\begin{document}

\title[Message Passing for Complex Question Answering]{Message Passing for Complex Question Answering\\ over Knowledge Graphs}

\author{Svitlana Vakulenko}
\affiliation{%
  \institution{Vienna University of\\
Economics and Business}
  \city{Vienna} 
  \country{Austria}
}
\email{svitlana.vakulenko@wu.ac.at}

\author{Javier David Fernandez Garcia}
\affiliation{%
  \institution{Vienna University of\\
Economics and Business}
  \city{Vienna} 
  \country{Austria}
}
\email{jfernand@wu.ac.at}

\author{Axel Polleres}
\affiliation{%
  \institution{Vienna University of\\
Economics and Business}
  \city{Vienna} 
  \country{Austria}
}
\email{axel.polleres@wu.ac.at}

\author{Maarten de Rijke}
\affiliation{%
  \institution{University of Amsterdam}
  \city{Amsterdam} 
  \country{The Netherlands}}
\email{derijke@uva.nl}

\author{Michael Cochez}
\affiliation{%
  \institution{Fraunhofer FIT}
  \city{Sankt Augustin} 
  \country{Germany}}
\email{michaelcochez@gmail.com}

\renewcommand{\shortauthors}{Vakulenko et al.}

\begin{abstract}
Question answering over knowledge graphs (KGQA) has evolved from simple single-fact questions to complex questions that require graph traversal and aggregation.
We propose a novel approach for complex KGQA that uses unsupervised message passing, which propagates confidence scores obtained by parsing an input question and matching terms in the knowledge graph to a set of possible answers.
First, we identify entity, relationship, and class names mentioned in a natural language question, and map these to their counterparts in the graph.
Then, the confidence scores of these mappings propagate through the graph structure to locate the answer entities.
Finally, these are aggregated depending on the identified question type.

This approach can be efficiently implemented as a series of sparse matrix multiplications mimicking joins over small local subgraphs.
Our evaluation results show that the proposed approach outperforms the state-of-the-art on the LC-QuAD benchmark.
Moreover, we show that the performance of the approach depends only on the quality of the question interpretation results, i.e., given a correct relevance score distribution, our approach always produces a correct answer ranking. 
Our error analysis reveals correct answers missing from the benchmark dataset and inconsistencies in the DBpedia knowledge graph.
Finally, we provide a comprehensive evaluation of the proposed approach accompanied with an ablation study and an error analysis, which showcase the pitfalls for each of the question answering components in more detail.
\end{abstract}



%
% Keywords. The author(s) should pick words that accurately describe the work being
% presented. Separate the keywords with commas.

\maketitle

\section{Introduction}
The amount of data shared on the Web grows every day~\cite{gandomi2015beyond}.
Information retrieval systems are very efficient but they are limited in terms of the representation power for the underlying data structure that relies on an index for a single database table, i.e., a homogeneous collection of textual documents that share the same set of attributes, e.g., web pages or news articles~\cite{manning2010introduction}.
Knowledge graphs (KGs), i.e., graph-structured knowledge bases, such as DBpedia~\cite{dbpedia} or Wikidata~\cite{wikidata}, can interlink datasets with completely different schemas~\cite{DBLP:journals/dagstuhl-reports/BonattiDPP18}.
Moreover, SPARQL is a very expressive query language that allows us to retrieve data from a KG that matches specified graph patterns~\cite{sparql}.
Query formulation in SPARQL is not easy in practice since it requires knowledge of which datasets to access, their vocabulary and structure~\cite{DBLP:journals/internet/FreitasCOO12}.
Natural language interfaces can mitigate these issues, making data access more intuitive and also available for the majority of lay users~\cite{hendrix1982natural,DBLP:conf/semweb/KaufmannB07}.
One of the core functionalities for this kind of interfaces is question answering (QA), which goes beyond keyword or boolean queries, but also does not require knowledge of a specialised query language~\cite{DBLP:conf/rweb/UngerFC14}.

QA systems have been evolving since the early 1960s with early efforts in the database community to support natural language queries by translating them into structured queries~\citep[see, e.g.,][]{green-automatic-1963,woods-lunar-1977,bronnenberg-question-1980}.
Whereas a lot of recent work has considered answering questions using unstructured text corpora~\cite{DBLP:conf/emnlp/RajpurkarZLL16} or images~\cite{balanced_vqa_v2}, we consider the task of answering questions using information stored in KGs.
KGs are an important information source as an intermediate representation to integrate information from different sources and different modalities, such as images and text~\cite{DBLP:conf/www/FariaUSMF18}.
The resulting models are at the same time abstract, compact, and interpretable~\cite{wilcke2017knowledge}.

Question answering over knowledge graphs (KGQA) requires matching an input question to a subgraph, in the simplest case matching a single labeled edge (\emph{triple}) in the KG, a task also called \emph{simple} question answering~\cite{DBLP:journals/corr/BordesUCW15}.
The task of \emph{complex} question answering is defined in contrast to simple KGQA and requires matching more than one triple in the KG~\cite{DBLP:conf/semweb/TrivediMDL17}.
Previously proposed approaches to complex KGQA formulate it as a subgraph matching task~\cite{DBLP:conf/coling/BaoDYZZ16,maheshwari2018learning,DBLP:conf/coling/SorokinG18}, which is an NP-hard problem (by reduction to the subgraph isomorphism problem)~\cite{DBLP:conf/sigmod/ZouHWYHZ14}, or attempt to translate a natural language question into template-based SPARQL queries to retrieve the answer from the KG~\cite{DBLP:journals/corr/abs-1803-00832}, which requires a large number of candidate templates~\cite{DBLP:journals/corr/abs-1809-10044}.

% answering complex questions over KGs
We propose an approach to complex KGQA, called \OurApproach{}, based on an unsupervised message-passing algorithm, which allows for efficient reasoning under uncertainty using text similarity and the graph structure.
The results of our experimental evaluation demonstrate that \OurApproach{} is able to manage uncertainties in interpreting natural language questions, overcoming inconsistencies in a KG and incompleteness in the training data, conditions that restrict applications of alternative supervised approaches.

A core aspect of \OurApproach{} is in disentangling reasoning from the question interpretation process.
We show that uncertainty in reasoning stems from the question interpretation phase alone, meaning that under correct question interpretations \OurApproach{} will always rank the correct answers at the top.
\OurApproach{} is designed to accommodate uncertainty inherent in perception and interpretation processes via confidence scores that reflect natural language ambiguity, which %, just like for humans, 
depends on the ability to interpret terms correctly.
These ranked confidence values are then aggregated through our message-passing in a well-defined manner, which allows us to simultaneously consider multiple alternative interpretations of the seed terms, favoring the most likely interpretation in terms of the question context and relations modeled within the KG.
Rather than iterating over all possible orderings, we show how to evaluate multiple alternative question interpretations in parallel via efficient matrix operations.

Another assumption of \OurApproach{} that proves useful in practice is to deliberately disregard subject-object order, i.e., edge directions in a knowledge graph, thereby treating the graph as undirected.
Due to relation sparsity, this model relaxation turns out to be sufficient for most of the questions in the benchmark dataset.
We also demonstrate that due to insufficient relation coverage of the benchmark dataset any assumption on the correct order of the triples in the KG is prone to overfitting.
More than one question-answer example per relation is required to learn and evaluate a supervised model that predicts relation directionality.

Our evaluation on LC-QuAD\footnote{\url{http://lc-quad.sda.tech}}~\cite{DBLP:conf/semweb/TrivediMDL17}, a recent large-scale benchmark for complex KGQA, shows that \OurApproach{} significantly outperforms the state-of-the-art, without the need to translate a natural language question into a formal query language such as SPARQL.
We also show that \OurApproach{} is interpretable in terms of activation paths, and simple, effective and efficient at the same time. Moreover, our error analysis demonstrates limitations of the LC-QuAD benchmark, which was constructed using local graph patterns.

The rest of the paper is organized as follows. Section \ref{sec:related_work} summarizes the state of the art in KGQA. 
Section \ref{sec:approach} presents our approach, \OurApproach{}, with particular attention to the \emph{question interpretation} and \emph{answer inference} phases. 
In Section \ref{sec:evaluation}, we evaluate \OurApproach{} on the LC-QuAD dataset, providing a detailed ablation, scalability and error study. 
Finally, Section \ref{sec:conclusion} concludes and lists future work.
\section{Related Work}
\label{sec:related_work}
% \textbf{Question answering over knowledge graphs.}
% simple QA is solved
The most commonly used KGQA benchmark is the SimpleQuestions \cite{DBLP:journals/corr/BordesUCW15} dataset, which contains questions that require identifying a single triple to retrieve the correct answers.
Recent results~\cite{DBLP:conf/emnlp/PetrochukZ18} show that most of these simple questions can be solved using a standard neural network architectures.
This architecture consists of two components: (1) a conditional random fields (CRF) tagger with GloVe word embeddings for subject recognition given the text of the question, and (2) a bidirectional LSTM with FastText word embeddings for relation classification given the text of the question and the subject from the previous component.
Approaches to simple KGQA cannot easily be adapted to solving complex questions, since they rely heavily on the assumption that each question refers to only one entity and one relation in the KG, which is no longer the case for complex questions.
Moreover, complex KGQA also requires matching more complex graph patterns beyond a single triple.

Since developing KGQA systems requires solving several tasks, namely entity, relation and class linking, and afterwards query building, they are often implemented as independent components and arranged into a single pipeline~\cite{DBLP:conf/semweb/DubeyBCL18}.
Frameworks such as QALL-ME~\cite{DBLP:journals/ws/FerrandezSKDFNITONMG11}, OKBQA~\cite{DBLP:conf/semweb/KimUNFHKCKUKC17} and Frankenstein~\cite{DBLP:conf/esws/SinghBRS18}, allow one to share and reuse those components as a collaborative effort.
For example, Frankenstein includes 29 components that can be combined and interchanged~\cite{DBLP:conf/www/SinghRBSLUVKP0V18}.
However, the distribution of the number of components designed for each task is very unbalanced.
Most of the components in Frankenstein support entity and relation linking, 18 and 5 components respectively, while only two components perform query building~\cite{DBLP:journals/corr/abs-1809-10044}.

There is a lack of diversity in approaches that are being considered for retrieving answers from a KG.
OKBQA and Frankenstein both propose to translate natural language questions to SPARQL queries and then use existing query processing mechanism to retrieve answers.\footnote{\url{http://doc.okbqa.org/query-generation-module/v1/}}
We show that using matrix algebra approaches is more efficient in case of natural language processing than traditional SPARQL-based approaches since they are optimized for parallel computation, thereby allowing us to explore multiple alternative question interpretations at the same time~\cite{DBLP:conf/hpec/KepnerABBFGHKLM16,jamour2018demonstration}.

Query building approaches involve query generation and ranking steps~\cite{maheshwari2018learning,DBLP:conf/esws/ZafarNL18}.
These approaches essentially consider KGQA as a subgraph matching task~\cite{DBLP:conf/coling/BaoDYZZ16,maheshwari2018learning,DBLP:conf/coling/SorokinG18}, which is an NP-hard problem (by reduction to the subgraph isomorphism problem)~\cite{DBLP:conf/sigmod/ZouHWYHZ14}.
In practice, \citet{DBLP:journals/corr/abs-1809-10044} report that the question building components of Frankenstein fail to process 46\% questions from a subset of LC-QuAD due to the large number of triple patterns.
The reason is that most approaches to query generation are template-based~\cite{DBLP:journals/corr/abs-1803-00832} and complex questions require a large number of candidate templates~\cite{DBLP:journals/corr/abs-1809-10044}.
For example, WDAqua~\cite{DBLP:journals/corr/abs-1803-00832} generates 395  SPARQL queries as possible interpretations for the question ``Give me philosophers born in Saint Etienne.''

In summary, we identify the query building component as the main bottleneck for the development of KGQA systems and propose \OurApproach{} as an alternative to the query building approach.
Also, the pipeline paradigm is inefficient since it requires KG access first for disambiguation and then again for query building.
\OurApproach{} accesses the KG only to aggregate the confidence scores via graph traversal after question parsing and shallow linking that matches an input question to labels of nodes and edges in the KG.

The work most similar to ours is the spreading activation model of Treo~\cite{DBLP:journals/dke/FreitasOOSC13}, which is also a no-SPARQL approach based on graph traversal that propagates relatedness scores for ranking nodes with a cut-off threshold.
Treo relies on POS tags, the Stanford dependency parser, Wikipedia links and TF/IDF vectors for computing semantic relatedness scores between a question and terms in the KG.
Despite good performance on the QALD 2011 dataset, the main limitation of Treo is an average query execution time of 203s~\cite{DBLP:journals/dke/FreitasOOSC13}.
In this paper we show how to scale this kind of approach to large KGs and complement it with machine learning approaches for question parsing and word embeddings for semantic expansion.

Our approach overcomes the limitations of the previously proposed graph-based approach in terms of efficiency and scalability, which we demonstrate on a compelling benchmark.
We evaluate \OurApproach{} on LC-QuAD~\cite{DBLP:conf/semweb/TrivediMDL17}, which is the largest dataset used for benchmarking complex KGQA.
WDAqua is our baseline approach, which is the state-of-the-art in KGQA as the winner of the most recent Scalable Question Answering Challenge (SQA2018)~\cite{DBLP:conf/esws/NapolitanoUN18}.
Our evaluation results demonstrate improvements in precision and recall, while reducing average execution time over the SPARQL-based WDAqua, which is also orders of magnitude faster than results reported for the previous graph-based approach Treo.

There is other work on KGQA that uses embedding queries into a vector space~\cite{DBLP:conf/nips/HamiltonBZJL18,DBLP:conf/semweb/WangWLCZQ18}.
The benefit of our graph-based approach is in preserving the original structure of the KG that can be used for executing precise formal queries and answering ambiguous natural language questions.
The graph structure also makes the results traceable and, therefore, interpretable in terms of relevant paths and subgraphs in comparison with vector space operations.

\OurApproach{} uses message passing, a family of approaches that were initially developed in the context of probabilistic graphical models~\cite{pearl1988probabilistic,koller2009probabilistic}.
Graph neural networks trained to learn patterns of message passing have recently shown to be effective on a variety of tasks~\cite{DBLP:conf/icml/GilmerSRVD17,DBLP:journals/corr/abs-1806-01261}, including KG completion~\cite{DBLP:conf/esws/SchlichtkrullKB18}.
We show that our unsupervised message passing approach performs well on complex question answering and helps to overcome sampling biases in the training data, which supervised approaches are prone to.

\section{Approach}
\label{sec:approach}
\OurApproach{}, our KGQA approach, consists of two phases: (1) question interpretation, and (2) answer inference.
In the question interpretation phase we identify the sets of entities and predicates that we consider relevant for answering the input question along with the corresponding confidence scores.
In the second phase these confidence scores are propagated and aggregated directly over the structure of the KG, to provide a confidence distribution over the set of possible answers.
Our notion of KG is inspired by common concepts from the Resource Description Framework (RDF)~\cite{rdf-primer}, a standard representation used in many large-scale knowledge graphs, e.g., DBpedia and Wikidata:

\begin{definition}
\label{def:kg}
\rm
We define a (knowledge) \emph{graph} $K=\langle E,G,P\rangle$ as a tuple that contains sets of \emph{entities} $E$ (nodes) and \emph{properties} $P$, both represented by Unique Resource Identifiers (URIs), and a set of directed labeled edges $\langle e_i, p, e_j\rangle\in G$, where $e_i, e_j \in E$ and $p \in P$.
\end{definition}

\noindent%
The set of edges $G$ in a KG can be viewed as a (blank-node-free) RDF graph, with subject-predicate-object triples $\langle e_i, p, e_j\rangle$.
In analogy with RDFS, we refer to a subset of entities $C \subseteq E$ appearing as objects of the special property \texttt{\small rdf:type} as \emph{Classes}.
We also refer to classes, entities and properties collectively as \textit{terms}.
We ignore RDF literals, except for \texttt{\small rdfs:labels} that are used for matching questions to terms in KG.

The task of \emph{question answering over a knowledge graph} (KGQA) is: given a natural language question $Q$ and a knowledge graph $K$, produce the correct answer $A$, which is either a subset of entities in the KG $A \subseteq E$ or a result of a computation performed on this subset, such as the number of entities in this subset (\smalltt{COUNT}) or an assertion (\smalltt{ASK}).
These types of questions are the most frequent in existing KGQA benchmarks~\cite{DBLP:journals/corr/BordesUCW15,DBLP:conf/semweb/TrivediMDL17,DBLP:conf/semweb/UsbeckGN018}.
In the first phase \OurApproach{} maps a natural language question $Q$ to a structured model $q$, which the answer inference algorithm will operate on then. 

\subsection{Question interpretation}

To produce a question model $q$ we follow two steps: (1) \textit{parse}, which extracts references (entity, predicate and class mentions) from the natural language question and identifies the question type; and (2) \textit{match}, which assigns each of the extracted references to a ranked list of candidate entities, predicates and classes in the KG.

\begin{figure*}[!t]
\centering
\includegraphics[width=0.9\textwidth]{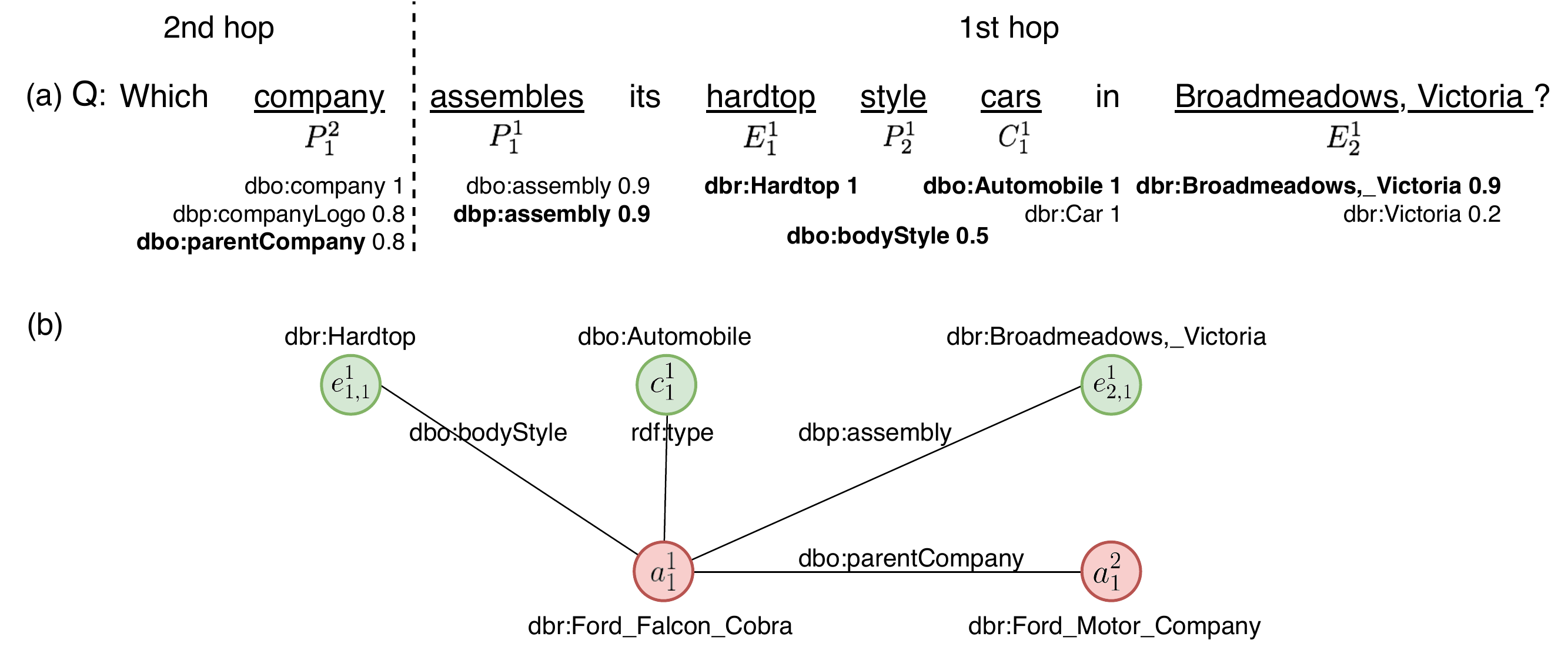}
\caption{(a) A sample question $Q$ highlighting different components of the question interpretation model: references and matched URIs with the corresponding confidence scores, along with (b) the illustration of a sample KG subgraph relevant to this question. The URIs in bold are the correct matches corresponding to the KG subgraph.}
\label{fig:qi}
\end{figure*}

Effectively, a complex question requires answering several sub-questions, which may depend on or support each other. A dependence relation between the sub-questions means that an answer $A^1$ to one of the questions is required to produce the answer $A^2$ for the other question: $A^2=f(A^1,K)$. 
We call such complex questions \textit{compound} questions and match the sequence in which these questions should be answered to \textit{hops} (in the context of this paper, one-variable graph patterns) in the KG.
Consider the sample compound question in Fig.~\ref{fig:qi}, which consists of two hops: (1) find the car types assembled in Broadmeadows Victoria, which have a hardtop style, (2) find the company, which produces these car types. There is an intermediate answer (the car types with the specified properties), which is required to arrive at the final answer (the company).

Accordingly, we define (compound) questions as follows:

\begin{definition}\label{def:question}
\rm
A \emph{question model} is a tuple $q=\langle t_q, Seq_q \rangle$, where $t_q \in T$ is a question type required to answer the question $Q$, and $Seq_q = (\langle E^i, P^i, C^i\rangle)_{i=1}^h$ is a sequence of $h$ hops over the KG, $E^i$ is a set of entity references, $P^i$ -- a set of property references, $C^i$ -- a set of class references relevant for the \textit{i}-hop in the graph, and $T$ -- a set of question types, such as $\{\smalltt{SELECT}, \smalltt{ASK}, \smalltt{COUNT}\}$.
\end{definition}

\noindent%
Hence, the question in Fig.~\ref{fig:qi} can be modeled as: $\langle \smalltt{SELECT}, (\langle E^1={}$\{``hardtop'', ``Broadmeadows, Victoria''\}, $P^1={}$\{``assembles'', ``style''\}, $C^1={}$\{``cars''\}$\rangle, \langle E^2=\emptyset, P^2={}$\{``company''\}, $C^2=\emptyset\rangle) \rangle$, where $E^i$, $P^i$, $C^i$ refer to the entities, predicates and classes in hop $i$.

Further, we describe how the question model $q$ is produced by parsing the input question $Q$, after which we match references in $q$ to entities and predicates in the graph $K$.

\OurParagraph{Parsing.}
Given a natural language question $Q$, the goal is to classify its type $t_q$ and parse it into a sequence $Seq_q$ of reference sets according to Definition~\ref{def:question}.
Question type detection is implemented as a supervised classification model trained on a dataset of annotated question-type pairs that learns to assign an input question to one of the predefined %question 
types $t_q \in T$.

We model reference (mention) extraction $Seq_q$ as a sequence labeling task~\cite{DBLP:conf/icml/LaffertyMP01}, in which a question is represented as a sequence of tokens (words or characters).
Then, a supervised machine learning model is trained on an annotated dataset to assign labels to tokens, which we use to extract references to entities, predicates and classes.
Moreover, we define the set of labels to group entities, properties and classes referenced in the question into $h$ hops.

\OurParagraph{Matching.}
Next, the question model (Definition~\ref{def:question}) is updated with an \emph{interpreted question model} $I(q) = (t_q, SEQ_q)$
%\todo[inline]{MC Check whether the following is correct}
%Here $SEQ_q$ has an entry for each step in $Seq_q$, containing possible matches in the knowledge graph and an assigned confidence.
%Each entry is a triple $\langle E^i, P^i, C^i \rangle$, where $X_i$ is a set of tuples which 
in which each component of $Seq_q$ is represented by sets of pairs from $(E \cup P \cup C) \times [0, 1]$ obtained by matching the references to concrete terms in $K$ (by their URIs) as follows: for each entity (or property, class, resp.) reference in $Seq_q$, we retrieve a ranked list of most similar entities from the KG along with the matching confidence score.

Fig.~\ref{fig:qi} also shows the result of this matching step on our example.
For instance, the property references for the first hop are replaced by the set of candidate URIs: $P^1 = \{P_1^1, P_2^1 \} \in SEQ_q$ within $I(q)$, where $P_1^1 = \{ \texttt{(\small  dbo:assembly}, 0.9)$, $(\texttt{\small dbp:assembly}, 0.9)\}$, $P_2^1 = \{ (\texttt{\small dbo:bodyStyle}, 0.5)\}$.

\subsection{Answer inference}

Our answer inference approach %is a recursive procedure that 
iteratively traverses and aggregates confidence scores across the graph based on the initial assignment from $I(q)$.
An answer set $A^i$, i.e., a set of entities along with their confidence scores $E \times [0, 1]$, is produced after each hop $i$ and used as part of the input to the next hop $i+1$, along with the %sets of 
terms matched for this hop in $I(q)$, i.e., $SEQ_q(i+1) = \langle E^{i+1}, P^{i+1}, C^{i+1}\rangle$.
The entity set $A^h$ produced after the last hop $h$ can be further transformed to produce the final answer: $A_q=f_{t_q}(A^h)$ via an aggregation function $f_{t_q} \in F$ from a predefined set of available aggregation functions $F$ defined for each of the question types $t_q \in T$.
We compute the answer set $A^i$ for each hop inductively in two steps: (1) \emph{subgraph extraction} and (2)~\emph{message passing}.

\OurParagraph{Subgraph extraction.} 
This step refers to the retrieval of relevant triples from the KG that form a subgraph.
Thus, the URIs of the matched entities and predicates in the query are used as seeds to retrieve the triples in the KG that contain at least one entity (in subject or object position), and one predicate from the corresponding reference sets.
Therefore, the extracted subgraph will contain $n$ entities, which include all entities from $E^{i}$ and the entities adjacent to them through properties from $P^{i}$.

The subgraph is represented as a set of $k$ adjacency matrices with $n$ entities in the subgraph: $\mathbb{S}^{k\times n\times n}$, where $k$ is the total number of matched property URIs.
There is a separate $n\times n$ matrix for each of the $k$ properties used as seeds, where $\mathbb{S}_{pij} = 1$ if there is an edge labeled $p$ between the entities $i$ and $j$ , and 0 otherwise.
All adjacency matrices are symmetric, because $I(q)$ does not model edge directionality, i.e., it treats $K$ as undirected.
Diagonal entries are assigned 0 to ignore self loops.

\begin{algorithm}[tb]
\caption{Message passing for KGQA}
\label{alg:mp}
\begin{flushleft}
\textbf{Input:} adjacency matrices of the subgraph $\mathbb{S}^{k\times n\times n}$,\\ entity $\mathbb{E}^{l\times n}$ and property reference activations $\mathbb{P}^{m\times k}$\\
\noindent\textbf{Output:} answer activations vector $A \in \mathbb{R}^n$\\ 
\end{flushleft}
\begin{algorithmic}[1]
\STATE  $W^{n}, N_P^{n}, \mathbb{Y_E}^{l\times n} = \emptyset$
\FOR{$P_j \in \mathbb{P}^{m\times k}, j \in \{1, ..., m\}$}%\commentA{iterate over property references}
    \STATE {$\mathbb{S}_j = \bigoplus_{i=1}^{k} P_j \otimes \mathbb{S}$} \commentA{property update}
    \STATE $\mathbb{Y} = \mathbb{E} \oplus\otimes \mathbb{S}_j$ \commentA{entity update}
    \STATE {$W = W + \bigoplus_{i=1}^{l} \mathbb{Y}_{ij}$} \commentA{sum of all activations}
    \STATE {${N}_{P_j} = \sum_{i=1}^{l} 1$ if $\mathbb{Y}_{ij} > 0$ else $0$} \STATE $\mathbb{Y_E} = \mathbb{Y_E} \oplus \mathbb{Y}$ \commentA{activation sums per entity}
\ENDFOR
\STATE $W = 2\cdot W / (l + m) $ \commentA{activation fraction}
\STATE $N_E = \sum_{i=1}^{l} 1$ if $\mathbb{Y_E}_{ij} > 0$ else $0$ \STATE \textbf{return} $A = (W \oplus N_E \oplus N_P) / (l + m + 1)$
\end{algorithmic}
\end{algorithm}

\begin{figure*}[!t]
\centering
\includegraphics[width=0.9\textwidth]{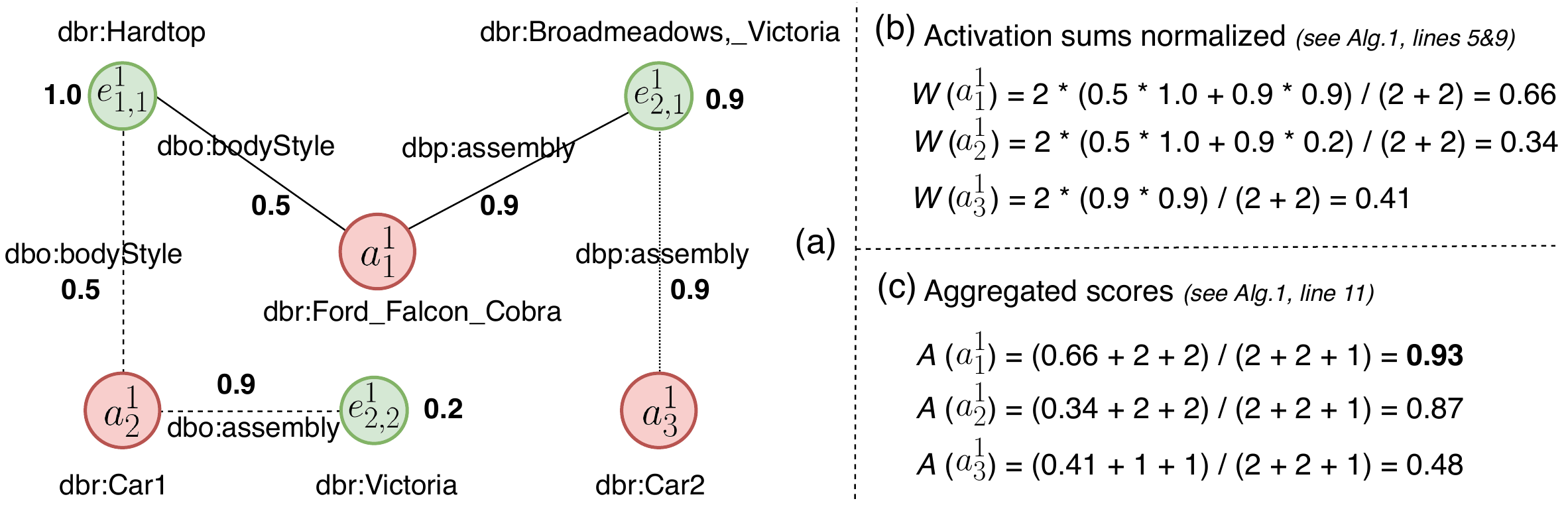}
 \caption{
(a) A sample subgraph with three entities as candidate answers, (b) their scores after predicate and entity propagation, and (c) the final aggregated score.
 }
\label{fig:agg}
\end{figure*}

\OurParagraph{Message passing.} The second step of the answer inference phase involves message passing,\footnote{The pseudocode of the message passing algorithm is presented in Algorithm~\ref{alg:mp}.} i.e., propagation of the confidence scores from the entities $E^i$ and predicates $P^i$, matched in the question interpretation phase, to adjacent entities in the extracted subgraph.
This process is performed in three steps, (1)~\emph{property update}, (2)~\emph{entity update}, and (3)~\emph{score aggregation}.
% See Alg. and the explanation below.
Algorithm~\ref{alg:mp} summarizes this process, detailed as follows.

For each of $m$ property references $P_j \in \mathbb{P}^{m\times k}, j \in \{1, \ldots, m\}$ where $m = |P^i|$, we
\begin{enumerate}[nosep,leftmargin=*]
\item \emph{select} the subset of adjacency matrices from $\mathbb{S}^{k\times n\times n}$ for the property URIs if $p_{ij} > 0$, where $p_{ij} \in P_j$, and propagate the confidence scores to the edges of the corresponding adjacency matrices via element-wise multiplication.
Then, all adjacency matrices are combined into a single adjacency matrix $\mathbb{S}_j^{n\times n}$, which contains all of their edges with the sum of confidence scores if edges overlap (\emph{property update}: line 3, Algorithm~\ref{alg:mp}).

\item \emph{perform} the main message-passing step via the sum-product update, in which the confidence scores from $l$ entity references, where $l = |E^i|$, are passed over to the adjacent entities via all edges in $\mathbb{S}_j^{n\times n}$ (\emph{entity update}: line 4, Algorithm~\ref{alg:mp}).

\item \emph{aggregate} the confidence scores for all $n$ entities in the subgraph into a single vector $A$ by combining the sum of all confidence scores with the number of entity and predicate reference sets, which received non-zero confidence score.
The intuition behind this \emph{score aggregation} formula (line 11,  Algorithm~\ref{alg:mp}) is that the answers that received confidence from the majority of entity and predicate references in the question should be preferred.
The computation of the answer scores for our running example is illustrated in Fig.~\ref{fig:agg}.
\end{enumerate}

\noindent%
The minimal confidence for the candidate answer is regulated by a threshold to exclude partial and low-confidence matches.
Finally, we also have an option to filter answers by considering only those entities in the answer set $A^i$ that have one of the classes in $C^i$.

The same procedure is repeated for each hop in the sequence using the corresponding URI activations for entities, properties and classes modeled in $SEQ_q(i)=\langle E^i, P^i, C^i\rangle$ and augmented with the intermediate answers produced for the previous hop $A^{i-1}$.
Lastly, the answer to the question $A_q$ is produced based on the entity set $A^h$, which is either returned `as is' or put through an aggregation function $f_{t_q}$ conditioned on the question type $t_q$.

\section{Evaluation Setup}
\label{sec:evaluation}

We evaluate \OurApproach{}, our KGQA approach, on the LC-QuAD dataset of complex questions constructed from the DBpedia KG~\cite{DBLP:conf/semweb/TrivediMDL17}.
% Firstly, we validate the correctness of the proposed message-passing algorithm using the gold-standard annotations by translating each SPARQL query into our question interpretation model.
First, we report the evaluation results of the end-to-end approach, which incorporates our message-passing algorithm in addition to the initial question interpretation (question parser and matching functions).
Second, we analyze the fraction and sources of errors produced by different KGQA components, which provides a comprehensive perspective on the limitations of the current state-of-the-art for KGQA, the complexity of the task, and limitations of the benchmark.
Our implementation and evaluation scripts are open-sourced.\footnote{\url{https://github.com/svakulenk0/KBQA}}

\OurParagraph{Baseline.}
We use WDAqua~\cite{DBLP:journals/corr/abs-1803-00832} as our baseline; to the best of our knowledge, the results produced by WDAqua are the only published results on the end-to-end question answering task for the LC-QuAD benchmark to date.
It is a rule-based framework that integrates several KGs in different languages and relies on a handful of SPARQL query patterns to generate SPARQL queries and rank them as likely question interpretations.
We rely on the evaluation results reported by the authors~\cite{DBLP:journals/corr/abs-1803-00832}.
WDAqua results were produced for the full LC-QuAD dataset, while other datasets were used for tuning the approach.

\OurParagraph{Metrics.}
We follow the standard evaluation metrics for the end-to-end KGQA task, i.e., we report precision (P) and recall (R) macro-averaged over all questions in the dataset, and then use them to compute the F-measure (F). 
Following the evaluation setup of the QALD-9 challenge~\cite{DBLP:conf/semweb/UsbeckGN018} we assign both precision and recall equal to 0 for every question in the following cases: (1) for \smalltt{SELECT} questions, no answer (empty answer set) is returned, while there is an answer (non-empty answer set) in the ground truth annotations; (2) for \smalltt{COUNT} or \smalltt{ASK} questions, an answer differs from the ground truth; (3) for all questions, the predicted answer type differs from the ground truth.
In the ablation study, we also analyze the fraction of questions with errors for each of the components separately, where an error is a not exact match with the ground-truth answer.

\OurParagraph{Hardware.}
We used a standard commodity server to train and evaluate \OurApproach{}: Intel(R) Core(TM) i7-8700K CPU @ 3.70GHz, RAM 16 GB DDR4 SDRAM 2400 MHz, 240 GB SSD, NVIDIA GP102 GeForce GTX 1080 Ti.

\subsection{The LC-QuAD dataset}
\label{sec:dataset}
The LC-QuAD dataset\footnote{\url{https://github.com/AskNowQA/LC-QuAD}}~\cite{DBLP:conf/semweb/TrivediMDL17} contains 5K question-query pairs that have correct answers in the DBpedia KG (2016-04 version).
The questions were generated using a set of SPARQL templates by seeding them with DBpedia entities and relations, and then paraphrased by human annotators.
All queries are of the form \smalltt{ASK}, \smalltt{SELECT}, and \smalltt{COUNT}, fit to subgraphs with diameter of at most 2-hops, contain 1--3 entities and 1--3 properties.

We used the train and test splits provided with the dataset (Table~\ref{tab:dataset}).
Two queries with no answers in the graph were excluded.
All questions are also annotated with ground-truth reference spans\footnote{\url{https://github.com/AskNowQA/EARL}} to evaluate performance of entity linking and relation detection~\cite{DBLP:conf/semweb/DubeyBCL18}.

\begin{table}
\centering
\caption{Dataset statistics: number of questions across the train and test splits; number of complex questions that reference more than one triple; number of complex questions that require two hops in the graph through an intermediate answer-entity.}
\label{tab:dataset}
\begin{tabular}{lrrr}
\toprule
 & \multicolumn{3}{c}{Questions} \\
\cmidrule{2-4}
Split & All & Complex & Compound  \\
\midrule
all  & 4,998 (100\%)       &     3,911 (78\%)       & 1,982 (40\%)           \\
train  & 3,999 \phantom{0}(80\%)       &      3,131 (78\%)           & 1,599 (40\%)          \\
test   & 999 \phantom{0}(20\%)        &       780 (78\%)          & 383 (38\%)         \\
\bottomrule
\end{tabular}
\end{table}

\subsection{Implementation details}
\label{sec:implementation}

Our implementation uses the English subset of the official DBpedia 2016-04 dump losslessly compressed into a single HDT file\footnote{\url{http://fragments.dbpedia.org/hdt/dbpedia2016-04en.hdt}}~\cite{FMPGPA:13}.
HDT is a state-of-the-art compressed RDF self-index, which scales linearly with the size of the graph and is, therefore, applicable to very large graphs in practice.
This KG contains 1B triples, more than 26M entities (\emph{dbpedia.org} namespace only) and 68,687 predicates.
Access to the KG for subgraph extraction and class constraint look-ups is implemented via the Python HDT API.\footnote{\url{https://github.com/Callidon/pyHDT}}
This API builds an additional index \cite{martinez2012exchange} to speed up all look-ups, and consumes the HDT mapped in disk, with a ${\sim}$3\% memory footprint.\footnote{Overall, DBpedia 2016-04 takes 18GB in disk, and 0.5GB in main memory.}

Our end-to-end KGQA solution integrates several components that can be trained and evaluated independently.
The pipeline includes two supervised neural networks for (1)~question type detection and (2)~reference extraction; and unsupervised functions for (3)~entity and (4)~predicate matching, and (5)~message passing.

\OurParagraph{Parsing.}
Question type detection is implemented as a bi-LSTM neural-network classifier trained on pairs of question and type.
We use another biLSTM+CRF neural network for extracting references to entities, classes and predicates for at most two hops using the set of six labels: \{``E1'', ``P1'', ``C1'', ``E2'', ``P2'', ``C2''\}.
Both classifiers use GloVe word embeddings pre-trained on the Common Crawl corpus with 840B tokens and 300 dimensions~\cite{pennington2014glove}. 

\OurParagraph{Matching.}
The labels of all entities and predicates in the KG (\texttt{\small rdfs: label} links) are indexed into two separate catalogs and embedded into two separate vector spaces using the English FastText model trained on Wikipedia~\cite{bojanowski2017enriching}.
We use two ranking functions for matching and assigning the corresponding confidence scores: index-based for entities and embedding-based for predicates.
The index-based ranking function uses BM25~\cite{manning2010introduction} to calculate confidence scores for the top-500 matches on the combination of n-grams and Snowball stems.\footnote{
    \url{https://www.elastic.co/guide/en/elasticsearch/reference/current}}
Embedding-based confidence scores are computed using the Magnitude library\footnote{\url{https://github.com/plasticityai/magnitude}}~\cite{patel2018magnitude} for the top-50 nearest neighbors in the FastText embedding space.

Many entity references in the LC-QuAD questions can be handled using simple string similarity matching techniques; e.g., `companies' can be mapped to ``http://dbpedia.org/ontology/Comp\-any''.
We built an ElasticSearch (Lucene) index to efficiently retrieve such entity candidates via string similarity to their labels.
The entity labels were automatically generated from entity URIs by stripping the domain part of the URI and lower-casing, e.g., entity ``http://dbpedia.org/ontology/Company'' received the label ``company'' to better match question words.
LC-QuAD questions also contain more complex paraphrases of the entity URIs that require semantic similarity computation beyond fuzzy string matching, such as ``movie'' refers to ``http://dbpedia.org/ontology/Film'', ``stockholder'' to ``http://dbpedia.org/property/owner'' or ``has kids'' to `http://dbpedia.org/ontology/child'.
We embeded entity and predicate labels with FastText~\cite{bojanowski2017enriching} to detect semantic similarities beyond string matching.

Index-based retrieval scales much better than nearest neighbour computation in the embedding space, which is a crucial requirement for the 26M entity catalog.
In our experiments, syntactic similarity was sufficient for entity matching in most of the cases, while property matching required capturing more semantic variations and greatly benefited from using pre-trained embeddings.

\section{Evaluation Results}
\label{sec:results}
% \todo{SV in progress}

Table~\ref{tab:results1} shows the performance of \OurApproach{} on the KGQA task in comparison with the results previously reported by  \citet{DBLP:journals/corr/abs-1803-00832}.
There is a noticeable improvement in recall (we were able to retrieve answers to 50\% of the benchmark questions), while maintaining a comparable precision score.
For the most recent QALD challenge the guidelines were updated to penalize systems that miss the correct answers, i.e., that are low in recall, which gives a clear signal of its importance for this task~\cite{DBLP:conf/semweb/UsbeckGN018}.
While it is often trivial for users to filter out a small number of incorrect answers that stem from interpretation ambiguity, it is much harder for users to recover missing correct answers.
Indeed, we showed that \OurApproach{} is able to identify correct answers that were missing even from the benchmark dataset since they were overlooked by the benchmark authors due to sampling bias.

\begin{table}
\centering
\caption{Evaluation results. (*) P of the WDAqua baseline is estimated from the reported precision of 0.59 for answered questions only. Runtime is reported in seconds per question as an average across all questions in the dataset. The distribution of runtimes for \OurApproach{} is Min: 0.01, Median: 0.67 Mean: 0.72, Max: 13.83}
\label{tab:results1}
\begin{tabular}{lcccc}
\toprule
\bf Approach & \bf P & \bf R & \bf F & \bf Runtime \\
\midrule
WDAqua & 0.22\smash{\rlap{*}} & 0.38 & 0.28    & 1.50 s/q \\
\OurApproach{} (our approach)      & 0.25 & 0.50 & 0.33    &   0.72 s/q  \\
\bottomrule
\end{tabular}
\end{table}

\subsection{Ablation study} 
Table~\ref{tab:results2} summarizes the results of our ablation study for different setups.
We report the fraction of all questions that have at least one answer that deviates from the ground truth (\emph{Total} column), questions with missing term matches (\emph{No match}) and other errors. 
Revised errors is the subset of other errors that were considered as true errors in the manual error analysis.

\begin{table*}[!t]
\caption{Ablation study results. (*) Question model results set the optimal performance for our approach assuming that the question interpretation is perfectly aligned with the ground-truth annotations. We then estimate additional (new) errors produced by each of the KGQA components separately. The experiments marked with {\bf GT} use the term URIs and question types extracted from the ground truth queries. {\bf GT span$^+$} uses spans from the ground-truth annotations and then corrects the distribution of the matched entities/properties to mimic correct question interpretation with a low-confidence tail with alternative matches. {\bf PR} (Parsed Results) stands for predictions made by question parsing and matching models (see Section \ref{sec:implementation}).}
\label{tab:results2}
%\vspace{0.2cm}
 \setlength{\tabcolsep}{5pt}
\centering\small
\begin{tabular}{ll|cccc|ccc|c|cc}
\hline
 &    \multirow{3}{*}{\bf Setup}              & \multicolumn{4}{c|}{\multirow{2}{*}{\bf Question interpretation}} &  \multirow{3}{*}{\bf P}  & \multirow{3}{*}{\bf R} & \multirow{3}{*}{\bf F}& \multicolumn{3}{c}{\bf Questions with errors}       \\
\cline{10-12}
   &    &  &  & &    &     &   &    & \multirow{2}{*}{Total}           & \multicolumn{2}{c}{New errors} \\
   \cline{3-6}\cline{11-12}
   &    & Q. type & Entity & Property & Class   &   &   &  &        & No match & Other$\rightarrow$Revised  \\
   
% \midrule
\hline
% \hline
1   & Question model*     & \multicolumn{4}{c|}{\cellcolor{lightgray}GT}            & 0.97 & 0.99 & 0.98 & \phantom{0}9\%             & --        & ~9\%   $\rightarrow$ ~5\%     \\
\hline
% \hline
2   & Question type  & PR & \multicolumn{3}{c|}{\cellcolor{lightgray} GT}    & 0.96 & 0.98 & 0.97 & 10\%             & --        & ~1\%   $\rightarrow$ ~1\%     \\
\hline
3   & Ignore classes  & \multicolumn{3}{c|}{\cellcolor{lightgray}GT}&    None      & 0.94 & 0.99 & 0.96 & 14\%             & --        & ~5\%   $\rightarrow$ ~3\%     \\
4   & Classes GT span$^+$  & \multicolumn{3}{c|}{\cellcolor{lightgray}GT}&    GT span$^+$     & 0.89 & 0.92 & 0.90 & 17\%             & \phantom{0}8\%      &  --    \\
\hline
5   & Entities GT span$^+$  & {\cellcolor{lightgray}GT}  &  GT span$^+$ & \multicolumn{2}{c|}{\cellcolor{lightgray} GT} & 0.85 & 0.88 & 0.86 & 20\%            & 10\%     & ~1\%   $\rightarrow$ ~1\%     \\
6   & Entities PR & {\cellcolor{lightgray}GT}& PR& \multicolumn{2}{c|}{\cellcolor{lightgray} GT} &  0.64 & 0.74 & 0.69 & 46\%            & 27\%     & 10\%  $\rightarrow$ ~5\%     \\
\hline
7   & Predicates GT span$^+$  & \multicolumn{2}{c}{\cellcolor{lightgray} GT} & GT span$^+$ & {\cellcolor{lightgray}GT}& 0.56 & 0.59 & 0.57 & 48\%            & 34\%     & ~5\%   $\rightarrow$ ~3\%     \\
8   & Predicates PR    & \multicolumn{2}{c}{\cellcolor{lightgray} GT} & PR & {\cellcolor{lightgray}GT} & 0.36 & 0.53 & 0.43 & 74\%            & 34\%     & 31\%  $\rightarrow$ 19\%    \\
\hline
\end{tabular}
\end{table*}

\begin{figure}[!t]
\centering
\includegraphics[width=0.37\textwidth]{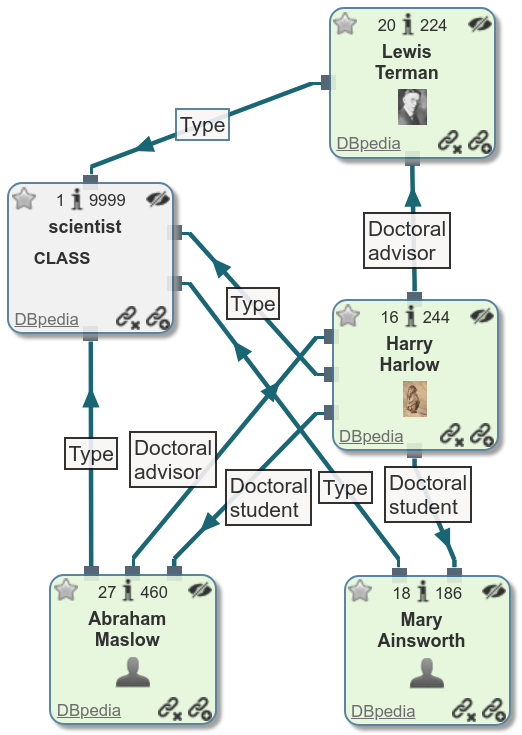}
\vspace{0.2cm}
\caption{\emph{Directed relation} example ({\tt\small dbp:doctoralStudents} and {\tt\small dbp:doctoralAdvisor} hierarchy) that requires modeling directionality of the relation. LC-QuAD question \#3267: ``Name the scientist whose supervisor also supervised Mary Ainsworth?'' (correct answer: Abraham Maslow) can be easily confused with a question: ``Name the scientist who supervised also the supervisor of Mary Ainsworth?'' (correct answer: Lewis Terman). LC-QuAD benchmark is not suitable for evaluating directionality interpretations, since only 35 questions (3.5\%) of the LC-QuAD test split use relations of this type, which explains high performance results of \OurApproach{} that treats all relation as undirected.}
\label{fig:supervision}
\end{figure}

\begin{figure}[!t]
\centering
\includegraphics[width=0.48\textwidth]{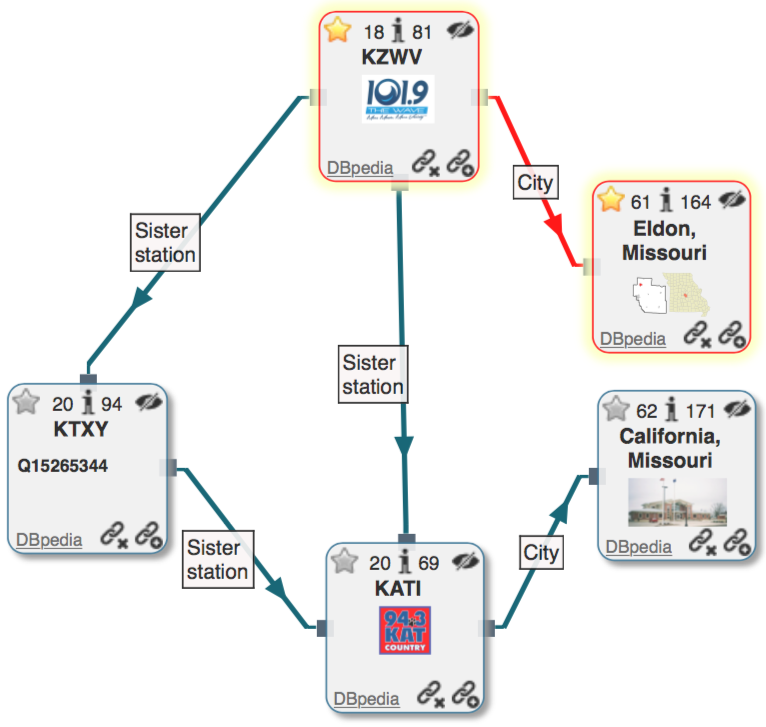}
\caption{\emph{Undirected relation} example ({\tt\small dbo:sisterStation}) that reflects bi-directional association between the adjacent entities (Missouri radio stations). LC-QuAD question \#4486: ``In which city is the sister station of KTXY located?'' (correct answer: {\tt\small dbr:California,Missouri}, {\tt\small dbr:Missouri}; missing answer: {\tt\small dbr:Eldon,Missouri}). DBpedia does not model bi-directional relations and the relation direction is selected at random in these cases. LC-QuAD does not reflect bi-directionality either by picking only one of the directions as the correct one and rejecting correct solutions ({\tt\small dbr:KZWY} $\rightarrow$ {\tt\small dbr:Eldon,Missouri}). \OurApproach{} was able to retrieve this false negative sample due to the default undirectionality assumption built into the question interpretation model.}
\label{fig:sister}
\end{figure}

Firstly, we make sure that the relaxations in our question interpretation model hold true for the majority of questions in the benchmark dataset (95\%)
by feeding all ground truth entity, class and property URIs to the answer inference module (Setup 1 in Table~\ref{tab:results2}).
We found that only 53 test questions (5\%) require one to model the exact order of entities in the triple, i.e., subject and predicate positions.
These questions explicitly refer to a hierarchy of entity relations, such as {\tt\small dbp:doctoralStudents} and {\tt\small dbp:doctoralAdvisor} (see Figure~\ref{fig:supervision}\footnote{The numbers on top of each entity show its number of predicates and triples.}$^,$\footnote{All sample graph visualizations illustrating different error types discovered in the LC-QuAD dataset in Figure~\ref{fig:supervision}, \ref{fig:sister}, \ref{fig:rome} were generated using the LODmilla web tool, \url{http://lodmilla.sztaki.hu}~\cite{sztaki8012}, with data from the DBpedia KG.}), and their directionality has to be interpreted to correctly answer such questions.
We also recovered a set of correct answers missing from the benchmark for  relations that are symmetric by nature, but were considered only in one direction by the benchmark, e.g., {\tt\small dbo:related}, {\tt\small dbo:associatedBand}, and {\tt\small dbo:sisterStation} (see Figure~\ref{fig:sister}).

These results indicate that a more complex question model attempting to reflect structural semantics of a question in terms of the expected edges and their directions (parse graph or lambda calculus) is likely to fall short when trained on this dataset: 53 sample questions are insufficient to train a reliable supervised model that can recognize relation directions from text, which explains poor results of a slot-matching model for subgraph ranking reported on this dataset~\cite{maheshwari2018learningSHORT}.

There were only 8 errors (1\%) due to the wrong question type detected caused by misspelled or grammatically incorrect questions (row 2 in Table~\ref{tab:results2}).
Next, we experimented with removing class constraints and found that although they generally help to filter out incorrect answers (row 3) our matching function missed many correct classes even using the ground-truth spans from the benchmark annotations (row 4).

The last four evaluation setups (5--8) in Table~\ref{tab:results2} show the errors from parsing and matching reference spans to entities and predicates in the KG.
Most errors were due to missing term matches (10--34\% of questions), which indicates that the parsing and matching functions constitute the bottleneck in our end-to-end KGQA.
Even with the ground-truth span annotations for predicate references the performance is below 0.6 (34\% of questions), which indicates that relation detection is much harder than the entity linking task, which is in line with results reported by \citet{DBLP:conf/semweb/DubeyBCL18} and \citet{DBLP:journals/corr/abs-1809-10044}.

The experiments marked \textbf{GT span+} were performed by matching terms to the KG using the ground-truth span annotations, then down-scaling the confidence scores for all matches and setting the confidence score of the match used in the ground-truth query to the maximum confidence score of 1.
In this setup, all correct answers according to the benchmark were ranked at the top, which demonstrates the correctness of the message passing and score aggregation algorithm.

\subsection{Scalability analysis}

\begin{figure}[!t]
\centering
\includegraphics[width=0.48\textwidth]{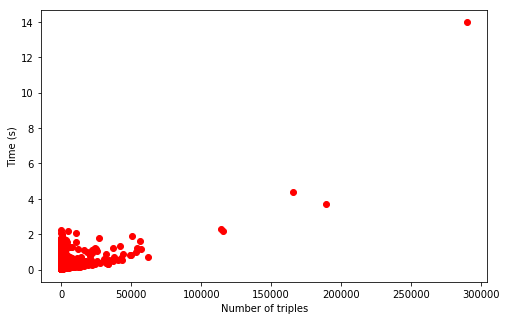}
\caption{Processing times per question from the LC-QuAD test split (Min: 0.01s Median: 0.68s Mean: 0.72s Max: 13.97s)}
\label{fig:scalability}
\end{figure}

As we reported in Table~\ref{tab:results1}, \OurApproach{} is twice as fast as  the WDAqua baseline using a comparable hardware configuration. 
Figure~\ref{fig:scalability} shows the distribution of processing times and the number of examined triples per question from the LC-QuAD test split. The results are in line with the expected fast retrieval of HDT ~\cite{FMPGPA:13}, which scales linearly with the size of the graph.
Most of the questions are processed within 2 seconds (with a median and mean around 0.7s), even those  examining more than 50K triples.
Note that only 10 questions took more than 2 seconds to process and 3 of them took more than 3 seconds. These outliers end up examining a large number of alternative interpretations (up to 300K triples), which could be prevented by setting a tighter threshold. Finally, it is worth mentioning that some questions end up with no results (i.e., 0 triples accessed), but they can take up to 2 seconds for parsing and matching. 
%On the other hand, parsing and matching can take up to 2 seconds without accessing the KG.

% \todo{rephrase: Motivated by our ablation study results, we implemented the fall-backs to consider all properties, when no property could be matched to a question, and ignoring answer class information altogether rather than relying on the matching results.}

% This result was achieved by retrieving at most 500 entity URIs with the confidence threshold at least 0.7 and 50 top-ranked predicate URIs per reference, and setting the confidence threshold for the answer set entities above 0.5.

% \todo{explain HDT caching @Javier}

\subsection{Error analysis}

We sampled questions with errors (P $< 1$ or R $< 1$) for each of the setups and performed an error analysis for a total of 206 questions.
Half of the errors were due to the incompleteness of the benchmark dataset and inconsistencies in the KG (column \emph{Revised} in Table~\ref{tab:results2}).
Since the benchmark provides only a single SPARQL query per question that contains a single URI for each entity, predicate and class, all alternative though correct matches are missing, e.g., the gold-truth query using {\tt\small dbp:writer} will miss {\tt\small dbo:writer}, or match all {\tt\small dbo:languages} triples but not {\tt\small dbo:language}, etc.

% \begin{figure}[!t]
% \centering
% \includegraphics[width=0.45\textwidth]{figs/devices.png}
% \caption{\emph{Alternative class} example that demonstrates a missing answer when only a single correct class URI is considered ({\tt\small dbr:information\_appliance} and not {\tt\small dbr:device}). LC-QuAD question \#733: ``Which technological products were manufactured by Foxconn?'' (correct answers: {\tt\small dbr:Amazon\_Kindle}, {\tt\small dbr:Xbox\_One}, {\tt\small dbr:PlayStation\_4}; missing answer: {\tt\small dbr:Fire\_Phone}). \OurApproach{} was able to retrieve this false negative sample due to the semantic matching function and retaining a list of alternative URIs per class mention.}
% \label{fig:devices}
% \end{figure}

\begin{figure}[!t]
\centering
\includegraphics[width=0.48\textwidth]{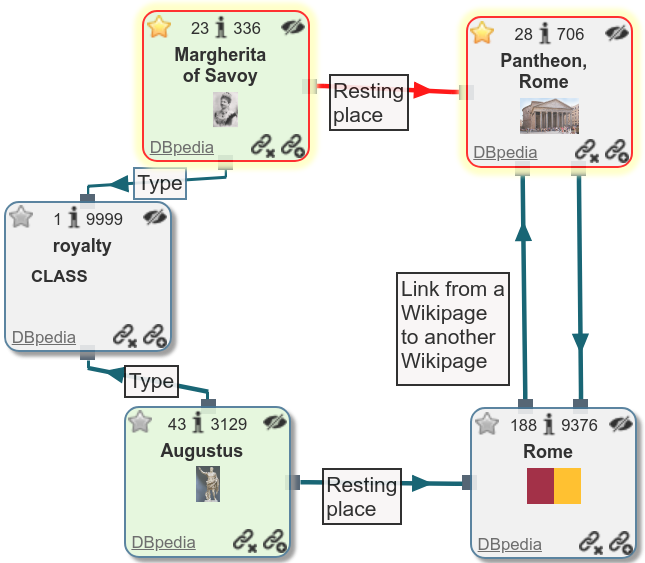}
%\vspace{0.2cm}
\caption{\emph{Alternative entity} example that demonstrates a missing answer when only a single correct entity URI is considered ({\tt\small dbr:Rome} and not {\tt\small dbr:Pantheon,Rome}).
LC-QuAD question \#261: ``Give me a count of royalties buried in Rome?'' (correct answer: {\tt\small dbr:Augustus}; missing answer: {\tt\small dbr:Margherita\_of\_Savoy}). \OurApproach{} was able to retrieve this false negative sample due to the string matching function and retaining a list of alternative URIs per entity mention.}
\label{fig:rome}
\end{figure}

\OurApproach{} was able to recover many such cases to produce additional correct answers using: (1) missing or alternative class URIs, e.g., {\tt\small dbr:Fire\_Phone} was missing from the answers for technological products manufactured by Foxconn since it was annotated as a {\tt\small device}, and not as an {\tt\small information appliance}; (2) related or alternative entity URIs, e.g., the set of royalties buried in {\tt\small dbo:Rome} should also include those buried in {\tt\small dbr:PantheonRome} (see Figure~\ref{fig:rome}); (3) alternative properties, e.g., {\tt\small dbo:hometown} as {\tt\small dbo:birthPlace}. 

% (see Figure~\ref{fig:devices})}

% \begin{figure}[!t]
% \centering
% \includegraphics[width=0.45\textwidth]{figs/race.png}
% \caption{\emph{Majority vote} example that demonstrates an effect of approximate reasoning that can retrieve alternative answers that do not fully match the input query. LC-QuAD question \#3302: ``Name the Pole driver of 1994 Spanish Grand Prix?'' (correct answer: {\tt\small dbr:Michael\_Schumacher}). \OurApproach{} also returns {\tt\small dbr:Ayrton\_Senna} as the second-best guess because he was the pole driver in many of the Grand Prix races. The model is not able to integrate other temporal facts that Senna died in an accident, while leading the 1994 San Marino Grand Prix, only a few weeks before the 1994 Spanish Grand Prix, and Michael Schumacher took over the first place in the rank afterwards.}
% \label{fig:race}
% \end{figure}

We discovered alternative answers due to the \textit{majority vote} effect, when many entities with low confidence help boost a single answer.
Majority voting can produce a best-effort guess based on the data in the KG even if the correct terms are missing from the KG or could not be recovered by the matching function, e.g., \emph{``In which time zone is Pong Pha?''} -- even if \emph{Pong Pha} is not in the KG many other locations with similar names are likely to be located in the same geographic area.
% However, this approach may also produce errors (see Figure~\ref{fig:race}).
% Such errors can be mitigated when differentiating between the activation patterns by adjusting the confidence scores with the number of evidence-source nodes, such that the candidate answers that received support from the top-match will be prefered over the candidates that received support from many nodes with a low confidence score.

Overall, our evaluation results indicate that the answer set of the LC-QuAD benchmark can be used only as a seed to estimate recall but does not provide us with a reliable metric for precision.
Attempts to further improve performance on such a dataset can lead to learning the biases embedded in the construction of the dataset, e.g., the set of relations and their directions.
\OurApproach{} is able to mitigate this pitfall by resorting to 
unsupervised message passing that collects answers from all local subgraphs containing terms matching the input question, in parallel.

\section{Conclusion}
\label{sec:conclusion}

We have proposed \OurApproach{}, a novel approach for complex KGQA using message passing, which sets the new state-of-the-art results on the LC-QuAD benchmark for complex question answering.
We have shown that \OurApproach{} is scalable and can be successfully applied to very large KGs, such as DBpedia, which is one of the biggest cross-domain KGs.
\OurApproach{} does not require supervision in the answer inference phase, which helps to avoid overfitting and to discover correct answers missing from the benchmark due to the limitations of its construction.
Moreover, the answer inference process can be explained by the extracted subgraph and the confidence score distribution.
\OurApproach{} requires only a handful of hyper-parameters to model confidence thresholds in order to stepwise filter partial results and trade off recall for precision.

% what we found;
\OurApproach{} is built on a basic assumption of considering edges as undirected in the graph, which proved reasonable and effective in our experiments.
The error analysis revealed that, in fact, symmetric edges were often missing in the KG, i.e., the decision on the order of entities in KG triples is made arbitrarily and is not duplicated in the reverse order.
However, there is also a (small) subset of relations, e.g., hierarchy relations, for which relation direction is essential.
% \todo[inline]{AP: IMHO, this next sentence could be thin ice, and while it is ok to mention it before in the evaluation, I wouldn't repeat/stress it in the conclusions, since this is essentially arguing that everything can be learned if enough data is available, sure, but in fact this directionality is maybe something where you rule-based heuristics in the parsing would actually work better.}
% We argue that these relations are underrepresented in the existing benchmark which will cause supervised approaches trained to detect relation direction to overfit the training data.

% what are the implications;
% \if0
Question answering over KGs is hard due to (1)~ambiguities stemming from question interpretation, (2)~inconsistencies in knowledge graphs, and (3)~challenges in constructing a reliable benchmark, which motivate the development of robust methods able to cope with uncertainties and also provide assistance to end-users in interpreting the provenance and confidence of the answers.
% we overcome the query building and showed very powerful maximize optimise for recall tune precision prune based on modeling query structure additional constraints on the walk 
% \fi

% what are the limitation;
\OurApproach{} is not without limitations. 
It is designed to handle questions where the answer is a subset of entities or an aggregate based on this subset, e.g., questions for which the expected answer is a subset of properties in the graph, are currently out of scope.
An important next step is to use \OurApproach{} to improve the recall of the benchmark dataset by complementing the answer set with missing answers derived from relaxing the dataset assumptions.
Recognizing relation directionality is an important direction for future work, which requires extending existing benchmark datasets and the addition of more cases where an explicit order is required to retrieve correct answers.
Another direction is to improve predicate matching, which is the weakest component of the proposed approach as identified in our ablation study.
Finally, unsupervised message passing can be adopted for other tasks that require uncertain reasoning on KGs, such as knowledge base completion, text entailment, summarization, and dialogue response generation.
% improve relation detection is the next bottleneck consider all increase recall implicit no lexical match

\subsection*{Acknowledgments}
This work was supported by 
the EU H2020 programme under the MSCA-RISE agreement 645751 ({RISE\_BPM}), 
the Austrian Research Promotion Agency (FFG) under projects CommuniData (855407) and Cityspin (861213),
Ahold Delhaize,
the Association of Universities in the Netherlands (VSNU),
and
the Innovation Center for Artificial Intelligence (ICAI).
All content represents the opinion of the authors, which is not necessarily shared or endorsed by their respective employers and/or sponsors.

% and SPECIAL (731601)
\bibliographystyle{ACM-Reference-Format}
\bibliography{refs}


\begin{thebibliography}{53}

%%% the Web address.

\ifx \showCODEN    \undefined \def \showCODEN     #1{\unskip}     \fi
\ifx \showDOI      \undefined \def \showDOI       #1{#1}\fi
\ifx \showISBNx    \undefined \def \showISBNx     #1{\unskip}     \fi
\ifx \showISBNxiii \undefined \def \showISBNxiii  #1{\unskip}     \fi
\ifx \showISSN     \undefined \def \showISSN      #1{\unskip}     \fi
\ifx \showLCCN     \undefined \def \showLCCN      #1{\unskip}     \fi
\ifx \shownote     \undefined \def \shownote      #1{#1}          \fi
\ifx \showarticletitle \undefined \def \showarticletitle #1{#1}   \fi
\ifx \showURL      \undefined \def \showURL       {\relax}        \fi
% invisible to TeX
\providecommand\bibfield[2]{#2}
\providecommand\bibinfo[2]{#2}
\providecommand\natexlab[1]{#1}
\providecommand\showeprint[2][]{arXiv:#2}

\bibitem[\protect\citeauthoryear{Bao, Duan, Yan, Zhou, and Zhao}{Bao
  et~al\mbox{.}}{2016}]%
        {DBLP:conf/coling/BaoDYZZ16}
\bibfield{author}{\bibinfo{person}{Jun{-}Wei Bao}, \bibinfo{person}{Nan Duan},
  \bibinfo{person}{Zhao Yan}, \bibinfo{person}{Ming Zhou}, {and}
  \bibinfo{person}{Tiejun Zhao}.} \bibinfo{year}{2016}\natexlab{}.
\newblock \showarticletitle{Constraint-Based Question Answering with Knowledge
  Graph}. In \bibinfo{booktitle}{\emph{{COLING} 2016}}.
  \bibinfo{pages}{2503--2514}.
\newblock


\bibitem[\protect\citeauthoryear{Battaglia, Hamrick, et~al\mbox{.}}{Battaglia
  et~al\mbox{.}}{2018}]%
        {DBLP:journals/corr/abs-1806-01261}
\bibfield{author}{\bibinfo{person}{Peter~W. Battaglia},
  \bibinfo{person}{Jessica~B. Hamrick}, {et~al\mbox{.}}}
  \bibinfo{year}{2018}\natexlab{}.
\newblock \showarticletitle{Relational inductive biases, deep learning, and
  graph networks}.
\newblock \bibinfo{journal}{\emph{CoRR}}  \bibinfo{volume}{abs/1806.01261}
  (\bibinfo{year}{2018}).
\newblock
\showeprint[arxiv]{1806.01261}


\bibitem[\protect\citeauthoryear{Bojanowski, Grave, Joulin, and
  Mikolov}{Bojanowski et~al\mbox{.}}{2017}]%
        {bojanowski2017enriching}
\bibfield{author}{\bibinfo{person}{Piotr Bojanowski}, \bibinfo{person}{Edouard
  Grave}, \bibinfo{person}{Armand Joulin}, {and} \bibinfo{person}{Tomas
  Mikolov}.} \bibinfo{year}{2017}\natexlab{}.
\newblock \showarticletitle{Enriching Word Vectors with Subword Information}.
\newblock \bibinfo{journal}{\emph{Transactions of the Association for
  Computational Linguistics}}  \bibinfo{volume}{5} (\bibinfo{year}{2017}),
  \bibinfo{pages}{135--146}.
\newblock
\showISSN{2307-387X}


\bibitem[\protect\citeauthoryear{Bonatti, Decker, Polleres, and
  Presutti}{Bonatti et~al\mbox{.}}{2018}]%
        {DBLP:journals/dagstuhl-reports/BonattiDPP18}
\bibfield{author}{\bibinfo{person}{Piero~Andrea Bonatti},
  \bibinfo{person}{Stefan Decker}, \bibinfo{person}{Axel Polleres}, {and}
  \bibinfo{person}{Valentina Presutti}.} \bibinfo{year}{2018}\natexlab{}.
\newblock \showarticletitle{Knowledge Graphs: New Directions for Knowledge
  Representation on the Semantic Web (Dagstuhl Seminar 18371)}.
\newblock \bibinfo{journal}{\emph{Dagstuhl Reports}} \bibinfo{volume}{8},
  \bibinfo{number}{9} (\bibinfo{year}{2018}), \bibinfo{pages}{29--111}.
\newblock


\bibitem[\protect\citeauthoryear{Bordes, Usunier, Chopra, and Weston}{Bordes
  et~al\mbox{.}}{2015}]%
        {DBLP:journals/corr/BordesUCW15}
\bibfield{author}{\bibinfo{person}{Antoine Bordes}, \bibinfo{person}{Nicolas
  Usunier}, \bibinfo{person}{Sumit Chopra}, {and} \bibinfo{person}{Jason
  Weston}.} \bibinfo{year}{2015}\natexlab{}.
\newblock \showarticletitle{Large-scale Simple Question Answering with Memory
  Networks}.
\newblock \bibinfo{journal}{\emph{CoRR}}  \bibinfo{volume}{abs/1506.02075}
  (\bibinfo{year}{2015}).
\newblock
\showeprint[arxiv]{1506.02075}


\bibitem[\protect\citeauthoryear{Bronnenberg, Bunt, Landsbergen, Scha,
  Schoenmakers, and van Utteren}{Bronnenberg et~al\mbox{.}}{1980}]%
        {bronnenberg-question-1980}
\bibfield{author}{\bibinfo{person}{Wim Bronnenberg}, \bibinfo{person}{Harry
  Bunt}, \bibinfo{person}{Jan Landsbergen}, \bibinfo{person}{Remko Scha},
  \bibinfo{person}{Wijnand Schoenmakers}, {and} \bibinfo{person}{Eric van
  Utteren}.} \bibinfo{year}{1980}\natexlab{}.
\newblock \showarticletitle{The question answering system {Phliqa1}}.
\newblock In \bibinfo{booktitle}{\emph{Natural Language Question Answering
  Systems}}, \bibfield{editor}{\bibinfo{person}{L.~Bolc}} (Ed.).
  \bibinfo{publisher}{MacMillan}, \bibinfo{pages}{217--305}.
\newblock


\bibitem[\protect\citeauthoryear{de~Faria, Usbeck, Sarullo, Mu, and
  Freitas}{de~Faria et~al\mbox{.}}{2018}]%
        {DBLP:conf/www/FariaUSMF18}
\bibfield{author}{\bibinfo{person}{Fabr{\'{\i}}cio~F. de Faria},
  \bibinfo{person}{Ricardo Usbeck}, \bibinfo{person}{Alessio Sarullo},
  \bibinfo{person}{Tingting Mu}, {and} \bibinfo{person}{Andr{\'{e}} Freitas}.}
  \bibinfo{year}{2018}\natexlab{}.
\newblock \showarticletitle{Question Answering Mediated by Visual Clues and
  Knowledge Graphs}. In \bibinfo{booktitle}{\emph{WWW}}.
  \bibinfo{pages}{1937--1939}.
\newblock


\bibitem[\protect\citeauthoryear{Diefenbach, Both, Singh, and Maret}{Diefenbach
  et~al\mbox{.}}{2018}]%
        {DBLP:journals/corr/abs-1803-00832}
\bibfield{author}{\bibinfo{person}{Dennis Diefenbach}, \bibinfo{person}{Andreas
  Both}, \bibinfo{person}{Kamal~Deep Singh}, {and} \bibinfo{person}{Pierre
  Maret}.} \bibinfo{year}{2018}\natexlab{}.
\newblock \showarticletitle{Towards a Question Answering System over the
  Semantic Web}.
\newblock \bibinfo{journal}{\emph{CoRR}}  \bibinfo{volume}{abs/1803.00832}
  (\bibinfo{year}{2018}).
\newblock
\showeprint[arxiv]{1803.00832}


\bibitem[\protect\citeauthoryear{Dubey, Banerjee, Chaudhuri, and Lehmann}{Dubey
  et~al\mbox{.}}{2018}]%
        {DBLP:conf/semweb/DubeyBCL18}
\bibfield{author}{\bibinfo{person}{Mohnish Dubey}, \bibinfo{person}{Debayan
  Banerjee}, \bibinfo{person}{Debanjan Chaudhuri}, {and} \bibinfo{person}{Jens
  Lehmann}.} \bibinfo{year}{2018}\natexlab{}.
\newblock \showarticletitle{{EARL:} Joint Entity and Relation Linking for
  Question Answering over Knowledge Graphs}. In \bibinfo{booktitle}{\emph{ISWC
  2018}}. \bibinfo{pages}{108--126}.
\newblock


\bibitem[\protect\citeauthoryear{Fern{\'a}ndez, Mart{\'\i}nez-Prieto,
  Guti{\'e}rrez, Polleres, and Arias}{Fern{\'a}ndez et~al\mbox{.}}{2013}]%
        {FMPGPA:13}
\bibfield{author}{\bibinfo{person}{Javier~D. Fern{\'a}ndez},
  \bibinfo{person}{Miguel~A. Mart{\'\i}nez-Prieto}, \bibinfo{person}{Claudio
  Guti{\'e}rrez}, \bibinfo{person}{Axel Polleres}, {and} \bibinfo{person}{Mario
  Arias}.} \bibinfo{year}{2013}\natexlab{}.
\newblock \showarticletitle{Binary RDF Representation for Publication and
  Exchange (HDT)}.
\newblock \bibinfo{journal}{\emph{J. Web Semant.}}  \bibinfo{volume}{19}
  (\bibinfo{year}{2013}), \bibinfo{pages}{22--41}.
\newblock


\bibitem[\protect\citeauthoryear{Ferr{\'{a}}ndez, Spurk, Kouylekov, Dornescu,
  Ferr{\'{a}}ndez, Negri, Izquierdo, Tom{\'{a}}s, Orasan, Neumann, Magnini, and
  Gonz{\'{a}}lez}{Ferr{\'{a}}ndez et~al\mbox{.}}{2011}]%
        {DBLP:journals/ws/FerrandezSKDFNITONMG11}
\bibfield{author}{\bibinfo{person}{{\'{O}}scar Ferr{\'{a}}ndez},
  \bibinfo{person}{Christian Spurk}, \bibinfo{person}{Milen Kouylekov},
  \bibinfo{person}{Iustin Dornescu}, \bibinfo{person}{Sergio Ferr{\'{a}}ndez},
  \bibinfo{person}{Matteo Negri}, \bibinfo{person}{Rub{\'{e}}n Izquierdo},
  \bibinfo{person}{David Tom{\'{a}}s}, \bibinfo{person}{Constantin Orasan},
  \bibinfo{person}{Guenter Neumann}, \bibinfo{person}{Bernardo Magnini}, {and}
  \bibinfo{person}{Jos{\'{e}} Luis~Vicedo Gonz{\'{a}}lez}.}
  \bibinfo{year}{2011}\natexlab{}.
\newblock \showarticletitle{The {QALL-ME} Framework: {A} specifiable-domain
  multilingual Question Answering architecture}.
\newblock \bibinfo{journal}{\emph{J. Web Semant.}} \bibinfo{volume}{9},
  \bibinfo{number}{2} (\bibinfo{year}{2011}), \bibinfo{pages}{137--145}.
\newblock


\bibitem[\protect\citeauthoryear{Freitas, Curry, Oliveira, and O'Riain}{Freitas
  et~al\mbox{.}}{2012}]%
        {DBLP:journals/internet/FreitasCOO12}
\bibfield{author}{\bibinfo{person}{Andr{\'{e}} Freitas},
  \bibinfo{person}{Edward Curry}, \bibinfo{person}{Jo{\~{a}}o~Gabriel
  Oliveira}, {and} \bibinfo{person}{Se{\'{a}}n O'Riain}.}
  \bibinfo{year}{2012}\natexlab{}.
\newblock \showarticletitle{Querying Heterogeneous Datasets on the Linked Data
  Web: Challenges, Approaches, and Trends}.
\newblock \bibinfo{journal}{\emph{{IEEE} Internet Computing}}
  \bibinfo{volume}{16}, \bibinfo{number}{1} (\bibinfo{year}{2012}),
  \bibinfo{pages}{24--33}.
\newblock


\bibitem[\protect\citeauthoryear{Freitas, Oliveira, O'Riain, da~Silva, and
  Curry}{Freitas et~al\mbox{.}}{2013}]%
        {DBLP:journals/dke/FreitasOOSC13}
\bibfield{author}{\bibinfo{person}{Andr{\'{e}} Freitas},
  \bibinfo{person}{Jo{\~{a}}o~Gabriel Oliveira}, \bibinfo{person}{Se{\'{a}}n
  O'Riain}, \bibinfo{person}{Jo{\~{a}}o Carlos~Pereira da Silva}, {and}
  \bibinfo{person}{Edward Curry}.} \bibinfo{year}{2013}\natexlab{}.
\newblock \showarticletitle{Querying linked data graphs using semantic
  relatedness: {A} vocabulary independent approach}.
\newblock \bibinfo{journal}{\emph{Data Knowl. Eng.}}  \bibinfo{volume}{88}
  (\bibinfo{year}{2013}), \bibinfo{pages}{126--141}.
\newblock


\bibitem[\protect\citeauthoryear{Gandomi and Haider}{Gandomi and
  Haider}{2015}]%
        {gandomi2015beyond}
\bibfield{author}{\bibinfo{person}{Amir Gandomi} {and} \bibinfo{person}{Murtaza
  Haider}.} \bibinfo{year}{2015}\natexlab{}.
\newblock \showarticletitle{Beyond the hype: Big data concepts, methods, and
  analytics}.
\newblock \bibinfo{journal}{\emph{International journal of information
  management}} \bibinfo{volume}{35}, \bibinfo{number}{2}
  (\bibinfo{year}{2015}), \bibinfo{pages}{137--144}.
\newblock


\bibitem[\protect\citeauthoryear{Gilmer, Schoenholz, Riley,
  et~al\mbox{.}}{Gilmer et~al\mbox{.}}{2017}]%
        {DBLP:conf/icml/GilmerSRVD17}
\bibfield{author}{\bibinfo{person}{Justin Gilmer}, \bibinfo{person}{Samuel~S.
  Schoenholz}, \bibinfo{person}{Patrick~F. Riley}, {et~al\mbox{.}}}
  \bibinfo{year}{2017}\natexlab{}.
\newblock \showarticletitle{Neural Message Passing for Quantum Chemistry}. In
  \bibinfo{booktitle}{\emph{ICML 2017}}. \bibinfo{pages}{1263--1272}.
\newblock


\bibitem[\protect\citeauthoryear{Goyal, Khot, et~al\mbox{.}}{Goyal
  et~al\mbox{.}}{2017}]%
        {balanced_vqa_v2}
\bibfield{author}{\bibinfo{person}{Yash Goyal}, \bibinfo{person}{Tejas Khot},
  {et~al\mbox{.}}} \bibinfo{year}{2017}\natexlab{}.
\newblock \showarticletitle{Making the {V} in {VQA} Matter: Elevating the Role
  of Image Understanding in {V}isual {Q}uestion {A}nswering}. In
  \bibinfo{booktitle}{\emph{CVPR 2017}}.
\newblock


\bibitem[\protect\citeauthoryear{Green, Wolf, Chomsky, and Laughery}{Green
  et~al\mbox{.}}{1963}]%
        {green-automatic-1963}
\bibfield{author}{\bibinfo{person}{Bert~F. Green}, \bibinfo{person}{Alice~K.
  Wolf}, \bibinfo{person}{Carol Chomsky}, {and} \bibinfo{person}{Kenneth
  Laughery}.} \bibinfo{year}{1963}\natexlab{}.
\newblock \showarticletitle{Baseball: An automatic question answerer.}
\newblock In \bibinfo{booktitle}{\emph{Computers and Thought}}.
  \bibinfo{publisher}{McGraw-Hill}, \bibinfo{pages}{219--224}.
\newblock


\bibitem[\protect\citeauthoryear{Hamilton, Bajaj, Zitnik, Jurafsky, and
  Leskovec}{Hamilton et~al\mbox{.}}{2018}]%
        {DBLP:conf/nips/HamiltonBZJL18}
\bibfield{author}{\bibinfo{person}{William~L. Hamilton}, \bibinfo{person}{Payal
  Bajaj}, \bibinfo{person}{Marinka Zitnik}, \bibinfo{person}{Dan Jurafsky},
  {and} \bibinfo{person}{Jure Leskovec}.} \bibinfo{year}{2018}\natexlab{}.
\newblock \showarticletitle{Embedding Logical Queries on Knowledge Graphs}. In
  \bibinfo{booktitle}{\emph{NeurIPS}}. \bibinfo{pages}{2030--2041}.
\newblock


\bibitem[\protect\citeauthoryear{Harris and Seaborne}{Harris and
  Seaborne}{2013}]%
        {sparql}
\bibfield{author}{\bibinfo{person}{Steve Harris} {and} \bibinfo{person}{Andy
  Seaborne}.} \bibinfo{year}{2013}\natexlab{}.
\newblock \bibinfo{title}{{SPARQL} 1.1 Query Language}.
\newblock \bibinfo{howpublished}{W3C Recommendation}.   (\bibinfo{date}{March}
  \bibinfo{year}{2013}).
\newblock


\bibitem[\protect\citeauthoryear{Hendrix}{Hendrix}{1982}]%
        {hendrix1982natural}
\bibfield{author}{\bibinfo{person}{Gary~G Hendrix}.}
  \bibinfo{year}{1982}\natexlab{}.
\newblock \showarticletitle{Natural-language interface}.
\newblock \bibinfo{journal}{\emph{Computational Linguistics}}
  \bibinfo{volume}{8}, \bibinfo{number}{2} (\bibinfo{year}{1982}),
  \bibinfo{pages}{56--61}.
\newblock


\bibitem[\protect\citeauthoryear{Jamour, Abdelaziz, and Kalnis}{Jamour
  et~al\mbox{.}}{2018}]%
        {jamour2018demonstration}
\bibfield{author}{\bibinfo{person}{Fuad Jamour}, \bibinfo{person}{Ibrahim
  Abdelaziz}, {and} \bibinfo{person}{Panos Kalnis}.}
  \bibinfo{year}{2018}\natexlab{}.
\newblock \showarticletitle{A demonstration of MAGiQ: matrix algebra approach
  for solving RDF graph queries}.
\newblock \bibinfo{journal}{\emph{Proc. the VLDB Endowment}}
  \bibinfo{volume}{11}, \bibinfo{number}{12} (\bibinfo{year}{2018}),
  \bibinfo{pages}{1978--1981}.
\newblock


\bibitem[\protect\citeauthoryear{Kaufmann and Bernstein}{Kaufmann and
  Bernstein}{2007}]%
        {DBLP:conf/semweb/KaufmannB07}
\bibfield{author}{\bibinfo{person}{Esther Kaufmann} {and}
  \bibinfo{person}{Abraham Bernstein}.} \bibinfo{year}{2007}\natexlab{}.
\newblock \showarticletitle{How Useful Are Natural Language Interfaces to the
  Semantic Web for Casual End-Users?}. In \bibinfo{booktitle}{\emph{ISWC}}.
  \bibinfo{pages}{281--294}.
\newblock


\bibitem[\protect\citeauthoryear{Kepner, Aaltonen, Bader, Bulu{\c{c}},
  Franchetti, Gilbert, Hutchison, Kumar, Lumsdaine, Meyerhenke, McMillan, Yang,
  Owens, Zalewski, Mattson, and Moreira}{Kepner et~al\mbox{.}}{2016}]%
        {DBLP:conf/hpec/KepnerABBFGHKLM16}
\bibfield{author}{\bibinfo{person}{Jeremy Kepner}, \bibinfo{person}{Peter
  Aaltonen}, \bibinfo{person}{David~A. Bader}, \bibinfo{person}{Aydin
  Bulu{\c{c}}}, \bibinfo{person}{Franz Franchetti}, \bibinfo{person}{John~R.
  Gilbert}, \bibinfo{person}{Dylan Hutchison}, \bibinfo{person}{Manoj Kumar},
  \bibinfo{person}{Andrew Lumsdaine}, \bibinfo{person}{Henning Meyerhenke},
  \bibinfo{person}{Scott McMillan}, \bibinfo{person}{Carl Yang},
  \bibinfo{person}{John~D. Owens}, \bibinfo{person}{Marcin Zalewski},
  \bibinfo{person}{Timothy~G. Mattson}, {and} \bibinfo{person}{Jos{\'{e}}~E.
  Moreira}.} \bibinfo{year}{2016}\natexlab{}.
\newblock \showarticletitle{Mathematical foundations of the {GraphBLAS}}. In
  \bibinfo{booktitle}{\emph{HPEC}}. \bibinfo{pages}{1--9}.
\newblock


\bibitem[\protect\citeauthoryear{Kim, Unger, Ngomo, Freitas, Hahm, Kim, Choi,
  Kim, Usbeck, Kang, and Choi}{Kim et~al\mbox{.}}{2017}]%
        {DBLP:conf/semweb/KimUNFHKCKUKC17}
\bibfield{author}{\bibinfo{person}{Jin{-}Dong Kim}, \bibinfo{person}{Christina
  Unger}, \bibinfo{person}{Axel{-}Cyrille~Ngonga Ngomo},
  \bibinfo{person}{Andr{\'{e}} Freitas}, \bibinfo{person}{YoungGyun Hahm},
  \bibinfo{person}{Jiseong Kim}, \bibinfo{person}{Gyu{-}Hyun Choi},
  \bibinfo{person}{Jeonguk Kim}, \bibinfo{person}{Ricardo Usbeck},
  \bibinfo{person}{Myoung{-}Gu Kang}, {and} \bibinfo{person}{Key{-}Sun Choi}.}
  \bibinfo{year}{2017}\natexlab{}.
\newblock \showarticletitle{{OKBQA:} an Open Collaboration Framework for
  Development of Natural Language Question-Answering over Knowledge Bases}. In
  \bibinfo{booktitle}{\emph{ISWC}}.
\newblock


\bibitem[\protect\citeauthoryear{Koller, Friedman, and Bach}{Koller
  et~al\mbox{.}}{2009}]%
        {koller2009probabilistic}
\bibfield{author}{\bibinfo{person}{Daphne Koller}, \bibinfo{person}{Nir
  Friedman}, {and} \bibinfo{person}{Francis Bach}.}
  \bibinfo{year}{2009}\natexlab{}.
\newblock \bibinfo{booktitle}{\emph{Probabilistic Graphical Models: Principles
  and Techniques}}.
\newblock \bibinfo{publisher}{MIT press}.
\newblock


\bibitem[\protect\citeauthoryear{Lafferty, McCallum, and Pereira}{Lafferty
  et~al\mbox{.}}{2001}]%
        {DBLP:conf/icml/LaffertyMP01}
\bibfield{author}{\bibinfo{person}{John~D. Lafferty}, \bibinfo{person}{Andrew
  McCallum}, {and} \bibinfo{person}{Fernando C.~N. Pereira}.}
  \bibinfo{year}{2001}\natexlab{}.
\newblock \showarticletitle{Conditional Random Fields: Probabilistic Models for
  Segmenting and Labeling Sequence Data}. In \bibinfo{booktitle}{\emph{ICML
  2001}}. \bibinfo{pages}{282--289}.
\newblock


\bibitem[\protect\citeauthoryear{Lehmann, Isele, Jakob, Jentzsch, Kontokostas,
  et~al\mbox{.}}{Lehmann et~al\mbox{.}}{2015}]%
        {dbpedia}
\bibfield{author}{\bibinfo{person}{Jens Lehmann}, \bibinfo{person}{Robert
  Isele}, \bibinfo{person}{Max Jakob}, \bibinfo{person}{Anja Jentzsch},
  \bibinfo{person}{Dimitris Kontokostas}, {et~al\mbox{.}}}
  \bibinfo{year}{2015}\natexlab{}.
\newblock \showarticletitle{{DBpedia} - {A} large-scale, multilingual knowledge
  base extracted from Wikipedia}.
\newblock \bibinfo{journal}{\emph{Semantic Web}} \bibinfo{volume}{6},
  \bibinfo{number}{2} (\bibinfo{year}{2015}), \bibinfo{pages}{167--195}.
\newblock


\bibitem[\protect\citeauthoryear{Maheshwari, Trivedi, et~al\mbox{.}}{Maheshwari
  et~al\mbox{.}}{2018a}]%
        {maheshwari2018learningSHORT}
\bibfield{author}{\bibinfo{person}{Gaurav Maheshwari},
  \bibinfo{person}{Priyansh Trivedi}, {et~al\mbox{.}}}
  \bibinfo{year}{2018}\natexlab{a}.
\newblock \showarticletitle{Learning to Rank Query Graphs for Complex Question
  Answering over Knowledge Graphs}.
\newblock \bibinfo{journal}{\emph{arXiv preprint arXiv:1811.01118}}
  (\bibinfo{year}{2018}).
\newblock


\bibitem[\protect\citeauthoryear{Maheshwari, Trivedi, Lukovnikov, Chakraborty,
  Fischer, and Lehmann}{Maheshwari et~al\mbox{.}}{2018b}]%
        {maheshwari2018learning}
\bibfield{author}{\bibinfo{person}{Gaurav Maheshwari},
  \bibinfo{person}{Priyansh Trivedi}, \bibinfo{person}{Denis Lukovnikov},
  \bibinfo{person}{Nilesh Chakraborty}, \bibinfo{person}{Asja Fischer}, {and}
  \bibinfo{person}{Jens Lehmann}.} \bibinfo{year}{2018}\natexlab{b}.
\newblock \showarticletitle{Learning to Rank Query Graphs for Complex Question
  Answering over Knowledge Graphs}.
\newblock \bibinfo{journal}{\emph{arXiv preprint arXiv:1811.01118}}
  (\bibinfo{year}{2018}).
\newblock


\bibitem[\protect\citeauthoryear{Manning, Raghavan, and Sch{\"u}tze}{Manning
  et~al\mbox{.}}{2010}]%
        {manning2010introduction}
\bibfield{author}{\bibinfo{person}{Christopher Manning},
  \bibinfo{person}{Prabhakar Raghavan}, {and} \bibinfo{person}{Hinrich
  Sch{\"u}tze}.} \bibinfo{year}{2010}\natexlab{}.
\newblock \showarticletitle{Introduction to information retrieval}.
\newblock \bibinfo{journal}{\emph{Natural Language Engineering}}
  \bibinfo{volume}{16}, \bibinfo{number}{1} (\bibinfo{year}{2010}),
  \bibinfo{pages}{100--103}.
\newblock


\bibitem[\protect\citeauthoryear{Mart{\'\i}nez-Prieto, Gallego, and
  Fern{\'a}ndez}{Mart{\'\i}nez-Prieto et~al\mbox{.}}{2012}]%
        {martinez2012exchange}
\bibfield{author}{\bibinfo{person}{Miguel~A Mart{\'\i}nez-Prieto},
  \bibinfo{person}{Mario~Arias Gallego}, {and} \bibinfo{person}{Javier~D
  Fern{\'a}ndez}.} \bibinfo{year}{2012}\natexlab{}.
\newblock \showarticletitle{Exchange and consumption of huge RDF data}. In
  \bibinfo{booktitle}{\emph{ESWC}}. \bibinfo{pages}{437--452}.
\newblock


\bibitem[\protect\citeauthoryear{Micsik, Turbucz, and Gy{\"o}r{\"o}k}{Micsik
  et~al\mbox{.}}{2014}]%
        {sztaki8012}
\bibfield{author}{\bibinfo{person}{Andr{\'a}s Micsik},
  \bibinfo{person}{S{\'a}ndor Turbucz}, {and} \bibinfo{person}{Attila
  Gy{\"o}r{\"o}k}.} \bibinfo{year}{2014}\natexlab{}.
\newblock \showarticletitle{LODmilla: a Linked Data Browser for All}. In
  \bibinfo{booktitle}{\emph{Posters\&Demos SEMANTiCS 2014}},
  \bibfield{editor}{\bibinfo{person}{Sack Harald}, \bibinfo{person}{Filipowska
  Agata}, \bibinfo{person}{Lehmann Jens}, {and} \bibinfo{person}{Hellmann
  Sebastian}} (Eds.). \bibinfo{publisher}{CEUR-WS.org},
  \bibinfo{pages}{31--34}.
\newblock


\bibitem[\protect\citeauthoryear{Napolitano, Usbeck, and Ngomo}{Napolitano
  et~al\mbox{.}}{2018}]%
        {DBLP:conf/esws/NapolitanoUN18}
\bibfield{author}{\bibinfo{person}{Giulio Napolitano}, \bibinfo{person}{Ricardo
  Usbeck}, {and} \bibinfo{person}{Axel{-}Cyrille~Ngonga Ngomo}.}
  \bibinfo{year}{2018}\natexlab{}.
\newblock \showarticletitle{The Scalable Question Answering Over Linked Data
  {(SQA)} Challenge 2018}. In \bibinfo{booktitle}{\emph{SemWebEval Challenge at
  ESWC}}. \bibinfo{pages}{69--75}.
\newblock


\bibitem[\protect\citeauthoryear{Patel, Sands, Callison-Burch, and
  Apidianaki}{Patel et~al\mbox{.}}{2018}]%
        {patel2018magnitude}
\bibfield{author}{\bibinfo{person}{Ajay Patel}, \bibinfo{person}{Alexander
  Sands}, \bibinfo{person}{Chris Callison-Burch}, {and}
  \bibinfo{person}{Marianna Apidianaki}.} \bibinfo{year}{2018}\natexlab{}.
\newblock \showarticletitle{Magnitude: A Fast, Efficient Universal Vector
  Embedding Utility Package}. In \bibinfo{booktitle}{\emph{EMNLP 2018}}.
  \bibinfo{pages}{120--126}.
\newblock


\bibitem[\protect\citeauthoryear{Pearl}{Pearl}{1988}]%
        {pearl1988probabilistic}
\bibfield{author}{\bibinfo{person}{Judea Pearl}.}
  \bibinfo{year}{1988}\natexlab{}.
\newblock \bibinfo{booktitle}{\emph{Probabilistic Reasoning in Intelligent
  Systems: Networks of Plausible Inference}}.
\newblock \bibinfo{publisher}{Morgan Kaufmann Publishers Inc.}
\newblock


\bibitem[\protect\citeauthoryear{Pennington, Socher, and Manning}{Pennington
  et~al\mbox{.}}{2014}]%
        {pennington2014glove}
\bibfield{author}{\bibinfo{person}{Jeffrey Pennington},
  \bibinfo{person}{Richard Socher}, {and} \bibinfo{person}{Christopher~D.
  Manning}.} \bibinfo{year}{2014}\natexlab{}.
\newblock \showarticletitle{GloVe: Global Vectors for Word Representation}. In
  \bibinfo{booktitle}{\emph{EMNLP 2014}}. \bibinfo{pages}{1532--1543}.
\newblock


\bibitem[\protect\citeauthoryear{Petrochuk and Zettlemoyer}{Petrochuk and
  Zettlemoyer}{2018}]%
        {DBLP:conf/emnlp/PetrochukZ18}
\bibfield{author}{\bibinfo{person}{Michael Petrochuk} {and}
  \bibinfo{person}{Luke Zettlemoyer}.} \bibinfo{year}{2018}\natexlab{}.
\newblock \showarticletitle{SimpleQuestions Nearly Solved: {A} New Upperbound
  and Baseline Approach}. In \bibinfo{booktitle}{\emph{EMNLP 2018}}.
  \bibinfo{pages}{554--558}.
\newblock


\bibitem[\protect\citeauthoryear{Rajpurkar, Zhang, Lopyrev, and
  Liang}{Rajpurkar et~al\mbox{.}}{2016}]%
        {DBLP:conf/emnlp/RajpurkarZLL16}
\bibfield{author}{\bibinfo{person}{Pranav Rajpurkar}, \bibinfo{person}{Jian
  Zhang}, \bibinfo{person}{Konstantin Lopyrev}, {and} \bibinfo{person}{Percy
  Liang}.} \bibinfo{year}{2016}\natexlab{}.
\newblock \showarticletitle{SQuAD: 100, 000+ Questions for Machine
  Comprehension of Text}. In \bibinfo{booktitle}{\emph{EMNLP 2016}}.
  \bibinfo{pages}{2383--2392}.
\newblock


\bibitem[\protect\citeauthoryear{Schlichtkrull, Kipf, Bloem, van~den Berg,
  Titov, and Welling}{Schlichtkrull et~al\mbox{.}}{2018}]%
        {DBLP:conf/esws/SchlichtkrullKB18}
\bibfield{author}{\bibinfo{person}{Michael~Sejr Schlichtkrull},
  \bibinfo{person}{Thomas~N. Kipf}, \bibinfo{person}{Peter Bloem},
  \bibinfo{person}{Rianne van~den Berg}, \bibinfo{person}{Ivan Titov}, {and}
  \bibinfo{person}{Max Welling}.} \bibinfo{year}{2018}\natexlab{}.
\newblock In \bibinfo{booktitle}{\emph{ESWC}}. \bibinfo{pages}{593--607}.
\newblock


\bibitem[\protect\citeauthoryear{Schreiber and Raimond}{Schreiber and
  Raimond}{2014}]%
        {rdf-primer}
\bibfield{author}{\bibinfo{person}{Guus Schreiber} {and} \bibinfo{person}{Yves
  Raimond}.} \bibinfo{year}{2014}\natexlab{}.
\newblock \bibinfo{title}{{RDF} 1.1 Primer}.
\newblock \bibinfo{howpublished}{W3C Note}.   (\bibinfo{date}{June}
  \bibinfo{year}{2014}).
\newblock


\bibitem[\protect\citeauthoryear{Singh, Both, Radhakrishna, and
  Shekarpour}{Singh et~al\mbox{.}}{2018a}]%
        {DBLP:conf/esws/SinghBRS18}
\bibfield{author}{\bibinfo{person}{Kuldeep Singh}, \bibinfo{person}{Andreas
  Both}, \bibinfo{person}{Arun~Sethupat Radhakrishna}, {and}
  \bibinfo{person}{Saeedeh Shekarpour}.} \bibinfo{year}{2018}\natexlab{a}.
\newblock \showarticletitle{Frankenstein: {A} Platform Enabling Reuse of
  Question Answering Components}. In \bibinfo{booktitle}{\emph{ESWC}}.
  \bibinfo{pages}{624--638}.
\newblock


\bibitem[\protect\citeauthoryear{Singh, Lytra, Radhakrishna, Shekarpour, Vidal,
  and Lehmann}{Singh et~al\mbox{.}}{2018b}]%
        {DBLP:journals/corr/abs-1809-10044}
\bibfield{author}{\bibinfo{person}{Kuldeep Singh}, \bibinfo{person}{Ioanna
  Lytra}, \bibinfo{person}{Arun~Sethupat Radhakrishna},
  \bibinfo{person}{Saeedeh Shekarpour}, \bibinfo{person}{Maria{-}Esther Vidal},
  {and} \bibinfo{person}{Jens Lehmann}.} \bibinfo{year}{2018}\natexlab{b}.
\newblock \showarticletitle{No One is Perfect: Analysing the Performance of
  Question Answering Components over the DBpedia Knowledge Graph}.
\newblock \bibinfo{journal}{\emph{CoRR}}  \bibinfo{volume}{abs/1809.10044}
  (\bibinfo{year}{2018}).
\newblock


\bibitem[\protect\citeauthoryear{Singh, Radhakrishna, Both, Shekarpour, Lytra,
  Usbeck, Vyas, Khikmatullaev, Punjani, Lange, Vidal, Lehmann, and Auer}{Singh
  et~al\mbox{.}}{2018c}]%
        {DBLP:conf/www/SinghRBSLUVKP0V18}
\bibfield{author}{\bibinfo{person}{Kuldeep Singh},
  \bibinfo{person}{Arun~Sethupat Radhakrishna}, \bibinfo{person}{Andreas Both},
  \bibinfo{person}{Saeedeh Shekarpour}, \bibinfo{person}{Ioanna Lytra},
  \bibinfo{person}{Ricardo Usbeck}, \bibinfo{person}{Akhilesh Vyas},
  \bibinfo{person}{Akmal Khikmatullaev}, \bibinfo{person}{Dharmen Punjani},
  \bibinfo{person}{Christoph Lange}, \bibinfo{person}{Maria{-}Esther Vidal},
  \bibinfo{person}{Jens Lehmann}, {and} \bibinfo{person}{S{\"{o}}ren Auer}.}
  \bibinfo{year}{2018}\natexlab{c}.
\newblock \showarticletitle{Why Reinvent the Wheel: Let's Build Question
  Answering Systems Together}. In \bibinfo{booktitle}{\emph{WWW}}.
  \bibinfo{pages}{1247--1256}.
\newblock


\bibitem[\protect\citeauthoryear{Sorokin and Gurevych}{Sorokin and
  Gurevych}{2018}]%
        {DBLP:conf/coling/SorokinG18}
\bibfield{author}{\bibinfo{person}{Daniil Sorokin} {and} \bibinfo{person}{Iryna
  Gurevych}.} \bibinfo{year}{2018}\natexlab{}.
\newblock \showarticletitle{Modeling Semantics with Gated Graph Neural Networks
  for Knowledge Base Question Answering}. In \bibinfo{booktitle}{\emph{COLING
  2018}}. \bibinfo{pages}{3306--3317}.
\newblock


\bibitem[\protect\citeauthoryear{Trivedi, Maheshwari, Dubey, and
  Lehmann}{Trivedi et~al\mbox{.}}{2017}]%
        {DBLP:conf/semweb/TrivediMDL17}
\bibfield{author}{\bibinfo{person}{Priyansh Trivedi}, \bibinfo{person}{Gaurav
  Maheshwari}, \bibinfo{person}{Mohnish Dubey}, {and} \bibinfo{person}{Jens
  Lehmann}.} \bibinfo{year}{2017}\natexlab{}.
\newblock \showarticletitle{LC-QuAD: {A} Corpus for Complex Question Answering
  over Knowledge Graphs}. In \bibinfo{booktitle}{\emph{ISWC}}.
  \bibinfo{pages}{210--218}.
\newblock


\bibitem[\protect\citeauthoryear{Unger, Freitas, and Cimiano}{Unger
  et~al\mbox{.}}{2014}]%
        {DBLP:conf/rweb/UngerFC14}
\bibfield{author}{\bibinfo{person}{Christina Unger},
  \bibinfo{person}{Andr{\'{e}} Freitas}, {and} \bibinfo{person}{Philipp
  Cimiano}.} \bibinfo{year}{2014}\natexlab{}.
\newblock \showarticletitle{An Introduction to Question Answering over Linked
  Data}. In \bibinfo{booktitle}{\emph{Reasoning Web}}.
  \bibinfo{pages}{100--140}.
\newblock


\bibitem[\protect\citeauthoryear{Usbeck, Gusmita, Ngomo, and Saleem}{Usbeck
  et~al\mbox{.}}{2018}]%
        {DBLP:conf/semweb/UsbeckGN018}
\bibfield{author}{\bibinfo{person}{Ricardo Usbeck}, \bibinfo{person}{Ria~Hari
  Gusmita}, \bibinfo{person}{Axel{-}Cyrille~Ngonga Ngomo}, {and}
  \bibinfo{person}{Muhammad Saleem}.} \bibinfo{year}{2018}\natexlab{}.
\newblock \showarticletitle{9th Challenge on Question Answering over Linked
  Data}. In \bibinfo{booktitle}{\emph{QALD at ISWC}}. \bibinfo{pages}{58--64}.
\newblock


\bibitem[\protect\citeauthoryear{Vrandecic and Kr{\"{o}}tzsch}{Vrandecic and
  Kr{\"{o}}tzsch}{2014}]%
        {wikidata}
\bibfield{author}{\bibinfo{person}{Denny Vrandecic} {and}
  \bibinfo{person}{Markus Kr{\"{o}}tzsch}.} \bibinfo{year}{2014}\natexlab{}.
\newblock \showarticletitle{Wikidata: a free collaborative knowledgebase}.
\newblock \bibinfo{journal}{\emph{Commun. {ACM}}} \bibinfo{volume}{57},
  \bibinfo{number}{10} (\bibinfo{year}{2014}), \bibinfo{pages}{78--85}.
\newblock


\bibitem[\protect\citeauthoryear{Wang, Wang, Liu, Chen, Zhang, and Qi}{Wang
  et~al\mbox{.}}{2018}]%
        {DBLP:conf/semweb/WangWLCZQ18}
\bibfield{author}{\bibinfo{person}{Meng Wang}, \bibinfo{person}{Ruijie Wang},
  \bibinfo{person}{Jun Liu}, \bibinfo{person}{Yihe Chen}, \bibinfo{person}{Lei
  Zhang}, {and} \bibinfo{person}{Guilin Qi}.} \bibinfo{year}{2018}\natexlab{}.
\newblock \showarticletitle{Towards Empty Answers in {SPARQL:} Approximating
  Querying with {RDF} Embedding}. In \bibinfo{booktitle}{\emph{ISWC}}.
  \bibinfo{pages}{513--529}.
\newblock


\bibitem[\protect\citeauthoryear{Wilcke, Bloem, and De~Boer}{Wilcke
  et~al\mbox{.}}{2017}]%
        {wilcke2017knowledge}
\bibfield{author}{\bibinfo{person}{Xander Wilcke}, \bibinfo{person}{Peter
  Bloem}, {and} \bibinfo{person}{Victor De~Boer}.}
  \bibinfo{year}{2017}\natexlab{}.
\newblock \showarticletitle{The knowledge graph as the default data model for
  learning on heterogeneous knowledge}.
\newblock \bibinfo{journal}{\emph{Data Science}} (\bibinfo{year}{2017}),
  \bibinfo{pages}{1--19}.
\newblock


\bibitem[\protect\citeauthoryear{Woods}{Woods}{1977}]%
        {woods-lunar-1977}
\bibfield{author}{\bibinfo{person}{William~Aaron Woods}.}
  \bibinfo{year}{1977}\natexlab{}.
\newblock \showarticletitle{Lunar rocks in natural {English}: {Explorations} in
  natural language question answering}.
\newblock In \bibinfo{booktitle}{\emph{Linguistic Structures Processing}},
  \bibfield{editor}{\bibinfo{person}{A.~Zampoli}} (Ed.).
  \bibinfo{publisher}{Elsevier North-Holland}, \bibinfo{pages}{521--569}.
\newblock


\bibitem[\protect\citeauthoryear{Zafar, Napolitano, and Lehmann}{Zafar
  et~al\mbox{.}}{2018}]%
        {DBLP:conf/esws/ZafarNL18}
\bibfield{author}{\bibinfo{person}{Hamid Zafar}, \bibinfo{person}{Giulio
  Napolitano}, {and} \bibinfo{person}{Jens Lehmann}.}
  \bibinfo{year}{2018}\natexlab{}.
\newblock \showarticletitle{Formal Query Generation for Question Answering over
  Knowledge Bases}. In \bibinfo{booktitle}{\emph{ESWC}}.
  \bibinfo{pages}{714--728}.
\newblock


\bibitem[\protect\citeauthoryear{Zou, Huang, Wang, Yu, He, and Zhao}{Zou
  et~al\mbox{.}}{2014}]%
        {DBLP:conf/sigmod/ZouHWYHZ14}
\bibfield{author}{\bibinfo{person}{Lei Zou}, \bibinfo{person}{Ruizhe Huang},
  \bibinfo{person}{Haixun Wang}, \bibinfo{person}{Jeffrey~Xu Yu},
  \bibinfo{person}{Wenqiang He}, {and} \bibinfo{person}{Dongyan Zhao}.}
  \bibinfo{year}{2014}\natexlab{}.
\newblock \showarticletitle{Natural language question answering over {RDF:} a
  graph data driven approach}. In \bibinfo{booktitle}{\emph{SIGMOD 2014}}.
  \bibinfo{pages}{313--324}.
\newblock


\end{thebibliography}

\end{document}